\def\year{2021}\relax
\documentclass[letterpaper]{article} 
\usepackage{aaai21}  
\usepackage{times}  
\usepackage{helvet} 
\usepackage{courier}  
\usepackage[hyphens]{url}  
\usepackage{graphicx} 

\usepackage{algorithm}  
\usepackage{algpseudocode} 
\usepackage{footnote}
\usepackage{graphicx}
\usepackage{epstopdf}
\usepackage{subfig}
\usepackage{inputenc}
\usepackage{sectsty}
\usepackage{amsmath}
\usepackage{amssymb}
\usepackage{tabularx}
\usepackage{multirow}
\usepackage{makecell}
\usepackage[toc,page]{appendix}

\usepackage{natbib}  
\usepackage{caption} 
\nocopyright

\title{Distributed Learning with Low Communication Cost via Gradient Boosting Untrained Neural Network}
\author {
    Xiatian Zhang, \textsuperscript{\rm 1} 
    Xunshi He, \textsuperscript{\rm 2}
    Nan Wang, \textsuperscript{\rm 3}
    Rong Chen \textsuperscript{\rm 4}
        
}
\affiliations {
    \textsuperscript{\rm 1} xiatian.zhang@foxmail.com \quad
    \textsuperscript{\rm 2} xunshi.he@gmail.com\quad
    \textsuperscript{\rm 3} nan.wang129@foxmail.com\quad
    \textsuperscript{\rm 4} winnie.chenrong@foxmail.com
    
}
\begin{document}

\maketitle
\begin{abstract}
For high-dimensional data, there are huge communication costs for distributed GBDT because the communication volume of GBDT is related to the number of features. To overcome this problem, we propose a novel gradient boosting algorithm, the Gradient Boosting Untrained Neural Network(GBUN). GBUN ensembles the untrained randomly generated neural network that softly distributes data samples to multiple neuron outputs and dramatically reduces the communication costs for distributed learning. To avoid creating huge neural networks for high-dimensional data, we extend Simhash algorithm to mimic forward calculation of the neural network. Our experiments on multiple public datasets show that GBUN is as good as conventional GBDT in terms of prediction accuracy and much better than it in scaling property for distributed learning. Comparing to conventional GBDT varieties, GBUN speeds up the training process up to 13 times on the cluster with 64 machines, and up to 4614 times on the cluster with 100KB/s network bandwidth. Therefore, GBUN is not only an efficient distributed learning algorithm but also has great potentials for federated learning.

\end{abstract}

\section{Introduction}
Gradient Boosting Decision Tree (GBDT)~\cite{friedman2001greedy} is currently the most cutting-edge and most widely used algorithm for general machine learning problems. The GBDT algorithm gradually builds multiple trees, each trying to minimize the previous step's training error. In this way, the complex loss function can be reduced step by step to obtain the optimal solution. Xgboost~\cite{chen2016xgboost} and LightGBM~\cite{ke2017lightgbm},  as the famous varieties of GBDT, have improved accuracy and efficiency by some optimizations on both mathematics and engineering. Both Xgboost and LightGBM 
have become the fundamental tools for solving general machine learning problems and has achieved great success in real applications. 

However, the distributed GBDT algorithm has huge communication costs, limiting the application of the algorithm on large-scale high-dimensional machine learning tasks. As we all know, the decision tree algorithm's split point search should scan the possible split points of all features to estimate information gain. Therefore, in distributed learning, searching a split point should collect the data of all the possible split points from all computing workers. At the same time, the communication volume increases as the rise of features and possible feature values. That is a disaster for the distributed GBDT algorithm on high-dimensional data.

To solve the performance bottleneck of GBDT in distributed learning, we propose a new algorithm, Gradient Boosting Untrained (Neural) Network (GBUN). GBUN ensembles the untrained randomly generated neural networks, and each softly distributes data samples to multiple neuron outputs. GBUN reduces a large amount of communication volume to train the base predictors in distributed training. However, neural networks are not suitable for handling high-dimensional data problems. Since a large number of input neurons make the weight matrix huge,  the storage and calculation overhead of the model will also be significant. Inspired by the Simhash~\cite{sadowski2007simhash} algorithm, we use hash tricks to generate neural network weights on-the-fly corresponding to non-zero-value features. That makes GBUN hold none weight matrix in memory and fulfill the forward calculation for high-dimensional sparse data efficiently.

We implement the Python and Spark libraries of the GBUN algorithm. One supports stand-alone training, and another supports distributed learning. Our experiments on multiple public datasets show that GBUN is as good as conventional GBDT in terms of prediction accuracy and much better than it in scaling property for distributed learning. Comparing to conventional GBDT varieties, GBUN speeds up the training process up to 13 times on the cluster with 64 machines, and up to 4614 times on the cluster with 100KB/s network bandwidth. Consequently, GBUN is not only an efficient distributed learning algorithm but also has great potentials for federated learning~\cite{konen2016federated, Yang@2019}.

The rest of this paper is as follows: Section 2 introduces the basic algorithm and distributed learning principles of GBUN; Section 3 reviews the related works; Section 4 presents the on-the-fly forward calculation method of the neural network for high-dimensional sparse data; Section 5 summarizes experimental results; Finally, Section 6 is the conclusion.

\section{Related Works}
\subsection{Boosting Neural Network}
As far as we know, there have been some works combining boosting methods and neural networks to obtain better prediction accuracy. \cite{schwenk2000boostingnn} proposed a technique that ensembles neural networks by AdaBoost. \cite{martnezmuoz2019sequential} introduced an approach that sequentially trains a shallow neural network by the gradient boosting. More than ensembling neural networks by gradient boosting, GrwoNet~\cite{badirli2020gradient} algorithm also takes the previous round's top hidden layer output as part of the next round input, which further improves the prediction accuracy. Both \cite{Moghimi2016, liu2019gradient} used the gradient boosting to ensemble the CNN networks. The former is for the time series forecasting, and the latter is for the image classification. All these works are similar to GBUN in the model structure, but their goals are to improve the prediction accuracy of the neural network model by boosting methods. All these algorithms should train the neural networks. Since the communication overhead of distributed neural network training is also huge, these algorithms are unlikely to improve the scaling property for distributed learning.

\subsection{Distributed GBDT}
The communication costs of distributed GBUN only depend on the number of untrained neural network's output neurons, which is small in usual. In contrast, the communication costs of distributed GBDT is related to the possible split points of all features. Both Xgboost and LightGBM have employed the histogram searching mode in the distributed implementations to reduce the communication volume, but the communication costs are still related to the number of features. Moreover, LightGBM has introduced \emph{Exclusive Feature Bundling} mechanism to reduce the number of features in the training process that makes it can handle at most millions of features in our distributed experiment. LightGBM also has adopted voting paralleling method~\cite{DBLP:conf/nips/MengKWCYML16} to decrease the communication volume. However, the experiments in its literature~\cite{DBLP:conf/nips/MengKWCYML16} have only verified 1200-dimensional datasets. Besides, We believe that the voting paralleling method implies two restrictions. The first is that the data distribution on different computing nodes must be independent identically distributed. The second is that this method is not suitable for high dimensional sparse data. Otherwise, the local $K$ best candidate split points on different computing workers may vary and result in a degenerated voting result. Therefore, we do not take the voting paralleling LightGBM as the comparison algorithm in our experiments.

\subsection{Random Methods in Machine Learning}
For a long time, random methods have been widely used in machine learning and data mining. Random hyperplane projection~\cite{plan2014dimension} can be employed to reduce the dimension of high-dimensional data. The Simhash algorithm~\cite{sadowski2007simhash}, as the derivation of the random hyperplane projection, is also an efficient approximate nearest neighbor searching method. Ensemble machine learning algorithms usually use the bagging and column sampling methods to construct multiple sub-training data sets to train different base predictors. Random Forest~\cite {liaw2002classification} is the most famous algorithm that has integrated both two approaches. As far as we know, the first ensemble learning algorithm that uses the total random base predictor is the random decision trees(RDT) algorithm~\cite{fan2003random}, which constructs multiple decision trees total randomly then averages the predicting scores of all trees as the final predicting score for a given sample. The random decision trees' prediction accuracy is as good as the random forest, and its training speed is much faster. Besides, random decision trees can also share purely random base predictors among different learning targets, significantly improving the efficiency of multi-target learning tasks ~\cite{Zhang10multi-labelclassification}. To overcome the performance problem of RDT for distributed learning, ~\cite{Zhang15RandomDecisionHashing} proposed the random decision hashing(RDH) algorithm that uses Simhash instead of the decision tree to divide the data space and achieve the same accuracy of RDT. In comparison, GBUN ensembles the randomly generated neural network by gradient boosting to achieve the same prediction accuracy as GBDT and much better scaling property than GBDT for distributed learning.


\section{Gradient Boosting Untrained Neural Network}
\begin{figure}[h!]
    \centering
    \includegraphics[width=0.95\linewidth, trim = 40 155 80 20, clip]{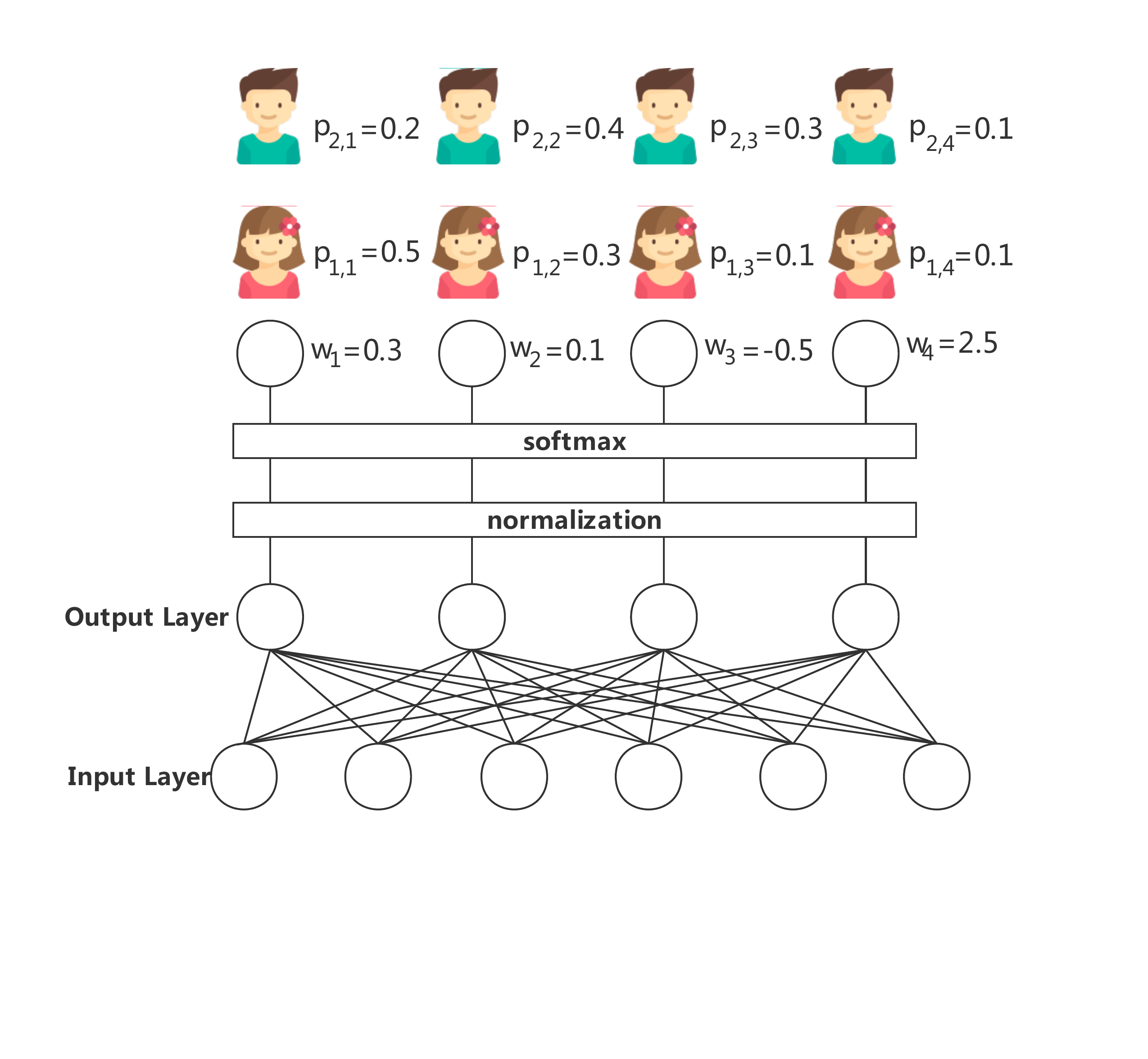}
    \caption{Untrained Neural network for softly splitting the data samples: $w_1, .., w_4$ are the predicting scores of the output neurons, $ p_ {1,1}, ..., p_ {1,4} $ and $ p_ {2,1}, ... , p_ {2,4} $ are the probabilities of two given samples assigned to the output neurons.}\label{fig:GBUN}
    \vspace{-2mm}
\end{figure}

Gradient boosting untrained neural networks, we refer to it as GBUN. For each boosting, GBUN randomly builds a single layer fully-connected neural network as a base predictor. Figure.~\ref{fig:GBUN} shows a simple example of the untrained neural network: A single-layer fully connected network with $ K $ outputs, a normalized layer, and a $softmax$ activation function. Referring to GBDT, the untrained neural network corresponds to the decision tree, and its output neurons are similar to the leaf nodes of the decision tree. However, GBUN doesn't assign one sample to a single output neuron. The $softmax$ function converts the output values to the probabilities that the sample belongs to each output neuron. Apparently, the neural network is also a data sample splitter like the decision tree but splitting softly.

The initialization of an untrained neural network is vital for the prediction accuracy of GBUN. But the popular initialization methods, such as Xavier~\cite{glorot2010init} and Kaiming~\cite{he2015delving} initialization, doesn't work well for GBUN. We adopt uniform distribution ranging -1 to 1 to generate the weight matrix based on some empirical study. We randomly set the 90\% weights in the matrix to 0 for better accuracy for the dense datasets.

Like the RDT\cite{Zhang10multi-labelclassification}, GBUN can share the randomly generated base predictors across all the learning targets for a given dataset. Consequently, no matter how many classes there are, GBUN only forwards the training samples once by the untrained network in each boosting for multi-class classification tasks. That makes GBUN have an advantage over GBDT on both model size and computational cost for the multi-class classification tasks.

\subsection{Basic GBUN Algorithm}
Based on the above introduction to the GBUN model, we will present the gradient boosting algorithm for the untrained neural network. 

We let the training dataset be $\mathcal{D}=\{(\mathbf{x}_i, y_i)\} (\left| \mathcal{D} \right|=n,\mathbf{x}_i \in \mathbb{R}^{m}, y_i \in \mathbb{R})$, which has $n$ samples and $m$ features. We also let the forward function of the untrained network in the $t$~round be $\mathcal{N}_t $, the number of output neurons be $K$, and the normalized operation be $norm$. Then, we let $\mathbf{p}_i =[p_{i1},p_{i2},...,p_{iK}], 0 \leq p_{ij} \leq 1$, $\sum_{j=1}^{K}p_{ij}=1, j=1,...,K$ , where $ p_ {ij} $ is the probability of the sample $\mathbf{x}_i$ assigned to the $j$-th output neuron of the untrained neural network. Besides, the predicting score vector is $\mathbf{W} = [w_1, w_2, ..., w_K]^{'}$,  where $w_j$ corresponds to the $j$-th output neuron. Afterward, we can derivate the prediction function $f_t(\mathbf{x}_i)$ corresponding to the $t$-th boosting model:
$$\mathbf{x}_i \xrightarrow{\ \ \mathcal{N}_t \ \ }  \mathbf{z}_i \xrightarrow{\ norm\ } \tilde{\mathbf{z}}_i \xrightarrow{softmax} \mathbf{p}_i \xrightarrow{<\cdot, \mathbf{W}>} f_t(\mathbf{x}_i)$$
Note that the $<>$ operator in the above equation is the inner product operation. 
According to above settings, we let the normalization equation be:
\begin{equation}\label{eq:norm} 
    norm(z_{ij})=\frac{z_{ij}-\overline{\mathbf{z}}_{\cdot j}}{\sigma(\mathbf{z}_{\cdot j})}
\end{equation}
where $z_{ij}$ is the output value of the $i$-th sample on the $j$-th output neuron. This normalization method is very similar to the classic batch normalization~\cite{ioffe2015batch} in deep learning.

In the following derivation process, we follow the setting of the objective function and gradient boosting framework of Xgboost. According to the Equation. (3) in the literature of Xgboost~\cite{chen2016xgboost}, the quadratic Taylor polynomial approximation of the residual objective function(originated from~\cite{Friedman98additivelogistic}) with the regularization term of $t$-th boosting is:

\begin{equation*}
\begin{split}
\text{obj}^{(t)} &\approx \sum_{i=1}^n [g_i f_t(x_i) + \frac{1}{2} h_i f_t(x_i)^2] +\gamma K + \frac{1}{2}\lambda \sum_{j=1}^K w_j^2 
\end{split}
\end{equation*}

We further expand the GBUN residual objective function into the following form:

\begin{equation}\label{eq:obj-our}
\begin{split}
\text{obj}^{(t)} \approx & \sum_{i=1}^n [g_i \sum_{j=1}^K p_{ij}w_j + \frac{1}{2} h_i (\sum_{j=1}^K p_{ij}w_j)^2]   + \frac{1}{2}\lambda \sum_{j=1}^K w_j^2
\end{split}
\end{equation}
In the above equation, because $K$ is fixed for a given $\mathcal{N}_t$, we omit $\gamma K$.

The Equation.~\eqref{eq:obj-our} is a quadratic function, and we can get the minimum solution by finding the point with 0 gradients. There is the derivative of  $obj^{(t)}$ with respect to $w_j$:

\begin{equation*}
\begin{split}
\frac{\partial \text{obj}^{(t)}}{\partial w_j} &\approx \sum_{i=1}^n [g_i p_{ij} +  h_i p_{ij}\sum_{k=1}^K p_{ik} w_k]  + \lambda w_j \\
& = \sum_{i=1}^n g_i p_{ij} + \sum_{k=1}^K w_k \sum_{i=1}^n h_i p_{ik} p_{ij}  + \lambda w_j
\end{split}
\end{equation*}
Then we let above formula be equal to 0, that is:
\begin{equation}\label{eq:diff-zero}
\sum_{i=1}^n g_i p_{ij} + \sum_{k=1}^K w_k \sum_{i=1}^n h_i p_{ik} p_{ij}  + \lambda w_j = 0
\end{equation}

Consequently, We have $K$ linear equations and $K$ variables. For simplicity, 

$$\mathbf{G}=[g_1,g_2,...,g_n]^{'}, \mathbf{H}=[h_1,h_2,...,h_n]^{'}$$
$$\mathbf{P}=\begin{bmatrix}
p_{11}  &  p_{12}  & \cdots & p_{1T}\\
p_{21} & p_{22}& \cdots & p_{2T} \\
\vdots & & \cdots & \vdots \\
p_{n1} & p_{n2} & \cdots & p_{nT}
\end{bmatrix}$$ 
$$\mathbf{H} \odot \mathbf{P} = \begin{bmatrix}
h_1p_{11}  &  h_1p_{12}  & \cdots & h_1p_{1T}\\
h_2p_{21} & h_2p_{22}& \cdots & h_2p_{2T} \\
\vdots & & \cdots & \vdots \\
h_np_{n1} & h_np_{n2} & \cdots & h_np_{nT}
\end{bmatrix}$$ 

In above equations, $\odot$ is the element-wise multiplication of a vector or matrix. When $\odot$ is in the middle of a vector and a matrix, it broadcasts the vector to the rows or columns of the matrix at first and then applies the element-wise multiplication. In addition, $\mathbf{I} $ is the identity matrix with rank $K$. Therefore:

\begin{equation*}
\begin{split}
    \mathbf{P}^{'} \mathbf{G} + \mathbf{P}^{'} (\mathbf{H} \odot \mathbf{P})\mathbf{W} + \lambda \mathbf{I}\mathbf{W} &= 0 \\
    [\mathbf{P}^{'} (\mathbf{H} \odot \mathbf{P})+\lambda \mathbf{I}]\mathbf{W} &= -\mathbf{P}^{'} \mathbf{G}
\end{split}
\end{equation*}
According to the method of solving the linear equations, the solution of the predicting scores can be obtained as follows,

\begin{equation}\label{eq:w-solve}
\begin{split}
    \mathbf{W} = -[\mathbf{P}^{'} (\mathbf{H} \odot \mathbf{P})+\lambda \mathbf{I}]^{-1}\mathbf{P}^{'}\mathbf{G}
\end{split}
\end{equation}

As shown from the above derivation, GBUN's learning method is simple and easy to implement. In the Algorithm.~\ref{alg:GBUN-basic}, there is the whole basic GBUN procedure. Note that, like the GBDT algorithm, we also introduce the shrinkage parameter~\cite{friedman2002stochastic} into the algorithm.

\begin{algorithm}[!h] 
    \caption{The Basic GBUN Algorithm}\label{alg:GBUN-basic}
    \begin{algorithmic}[1] 
      \State Train Data: $\mathbf{X}=[\mathbf{x}_1,...\mathbf{x}_i, ...\mathbf{x}_n], \mathbf{Y}=[y_1,...y_i, ...y_n]$
      \State \#$\mathcal{N}$ Outputs: $K$
      \State \#Rounds: $T$, Regulation Parameter: $\lambda$, Shrinkage: $\eta$
      \State Loss Function: $L(\hat{\mathbf{Y}}, \mathbf{Y})$
      \State $\hat{\mathbf{Y}}$ is predictions
      \Function{Training}{$L$, $\mathbf{X}$, $\mathbf{Y}$, $\lambda$, $\eta$, $\kappa$}
         \State Initialize \textbf{N\_list} to store $\mathcal{N}_t$
         \State Initialize \textbf{W\_list} to store $\mathbf{W_t}$
         \State $\mathbf{\hat{Y}} = \mathbf{0}$
         \For {$t$ = $1$ to $T$}
            \State Generate $\mathcal{N}_t$ with $m$ inputs and $K$ outputs.
            \State Put $\mathcal{N}_t$ into \textbf{N\_list}
            \State $\mathbf{G} \gets \partial L(\mathbf{\hat{Y}}, \mathbf{Y})/\partial \mathbf{\hat{Y}}$
            \State $\mathbf{H} \gets \partial^{2} L(\mathbf{\hat{Y}}, \mathbf{Y})/\partial\mathbf{\hat{Y}}^{2}$
            \State $\mathbf{Z} \gets \mathcal{N}(\mathbf{X})$
            \State $\tilde{\mathbf{Z}} \gets norm(\mathbf{Z})$
            \State $\mathbf{P} \gets softmax(\tilde{\mathbf{Z}})$
            \State $\mathbf{A} = \mathbf{P}^{'} (\mathbf{H} \odot \mathbf{P})+\lambda \mathbf{I}$
            \State $\mathbf{B} = -\mathbf{P}^{'}\mathbf{G}$
            \State $\mathbf{W}_t \gets -\mathbf{A}^{-1}\mathbf{B}$
            \State Put $\mathbf{W}_t$ into \textbf{W\_list}
            \State $\mathbf{\hat{Y}} \gets \mathbf{\hat{Y}}+\eta \mathbf{P}\mathbf{W}_t$
        \EndFor
        \State \Return \textbf{N\_list}, \textbf{W\_list}
      \EndFunction{}
    \end{algorithmic} 
\end{algorithm} 

\subsection{Distributed Learning}\label{sec:dist-learn-math}
In the previous subsection, we figured out the basic GBUN algorithm, and now we extend it to distributed learning. Assuming the dataset $\mathcal{D}=\{(\mathbf{x}_i, y_i)\} (\left| \mathcal{D} \right|=n,\mathbf{x}_i \in \mathbb{R}^{m}, y_i \in \mathbb{R})$ is divided into $S$ partitions, ie $\mathcal{D} = \cup_ {s = 1}^{S} \mathcal{D}^s$. Specifically, all terms with a subscript $s$ in the following equation indicate that it is related to data partition $\mathcal{D}^s$. In light of the Algorithm.~\ref{alg:GBUN-basic},  we can find that only normalization (step 16) and calculating the matrix $\mathbf{A}$ and $\mathbf{B}$(step 18 and 19) involve the whole training dataset. The rest of the steps can be done on each computing node. 

Firstly, in order to get normalized forward outputs, we calculate the following statistics for each $j$-th output neuron of the untrained neural network on each data partition $\mathcal{D}^s$:

\begin{equation} \label{eq:abc}
    a_{j}^s = \sum_{i=1}^{n^s} z_{ij}^s,\ \ \ \ \ \   b_{j}^s=\sum_{i=1}^{n^s} {z_{ij}^s}^2,\ \ \ \ \ \  n^s=|\mathcal{D}^s|
\end{equation}

Next, the global $\tilde{\mathbf{Z}}_{j}$ and $\sigma(\mathbf{Z}_{j})$ can be calculated as follows:
\begin{equation*}
\tilde{\mathbf{Z}}_{j} = \frac{\sum_{s=1}^S a_{j}^s}{\sum_{s=1}^S n^s}
, \quad
\sigma(\mathbf{Z}_{j}) = \sqrt{\frac{1}{\sum_{s=1}^S n^s - 1}(\frac{\sum_{s=1}^S b_{j}^s}{\sum_{s=1}^S n^s}-\tilde{\mathbf{Z}}_{j}^2)}
\end{equation*}
Then we can normalize the forward outputs locally by Equation.~\ref{eq:norm}.

Secondly, we introduce how to calculate the matrix $\mathbf{A}$ and $\mathbf{B}$ globally. Referring to the Equation.~\ref{eq:diff-zero}, we can find out that the elements in $\mathbf{A}$ and $\mathbf{B}$ are:

\begin{equation*}\label{eq:A}
\begin{split}
A_{kj} & = \sum_{i=1}^n h_i p_{ik} p_{ij} = \sum_{s=1}^S \sum_{i=1}^{n^s} h_{i}^s p_{ik}^s p_{ij}^s = \sum_{s=1}^S A_{kj}^s \\
B_{j} & = \sum_{i=1}^n g_i p_{ij} = \sum_{s=1}^S \sum_{i=1}^{n^s} g_{i}^s p_{ij}^s = \sum_{s=1}^S B_{j}^s
\end{split}
\end{equation*}
Then:
\begin{equation}
\mathbf{A}=\sum_{s=1}^S \mathbf{A}^s, \quad \mathbf{B}=\sum_{s=1}^S \mathbf{B}^s
\end{equation}

In summary, the distributed GBUN algorithm only performs two global operations in each boosting, and the communication costs of both above global operations are exclusively related to $K$, the number of outputs of the untrained network. Moreover, the $K$ only ranges from tens to about one thousand in practice. That makes the communication costs of distributed GBUN are much lower than distributed GBDT algorithms. 

\section{Forwarding High-dimensional Sparse Data}
\begin{figure}[!h]
    \centering
    \includegraphics[width=1.0\linewidth, trim=40 400 40 20,clip]{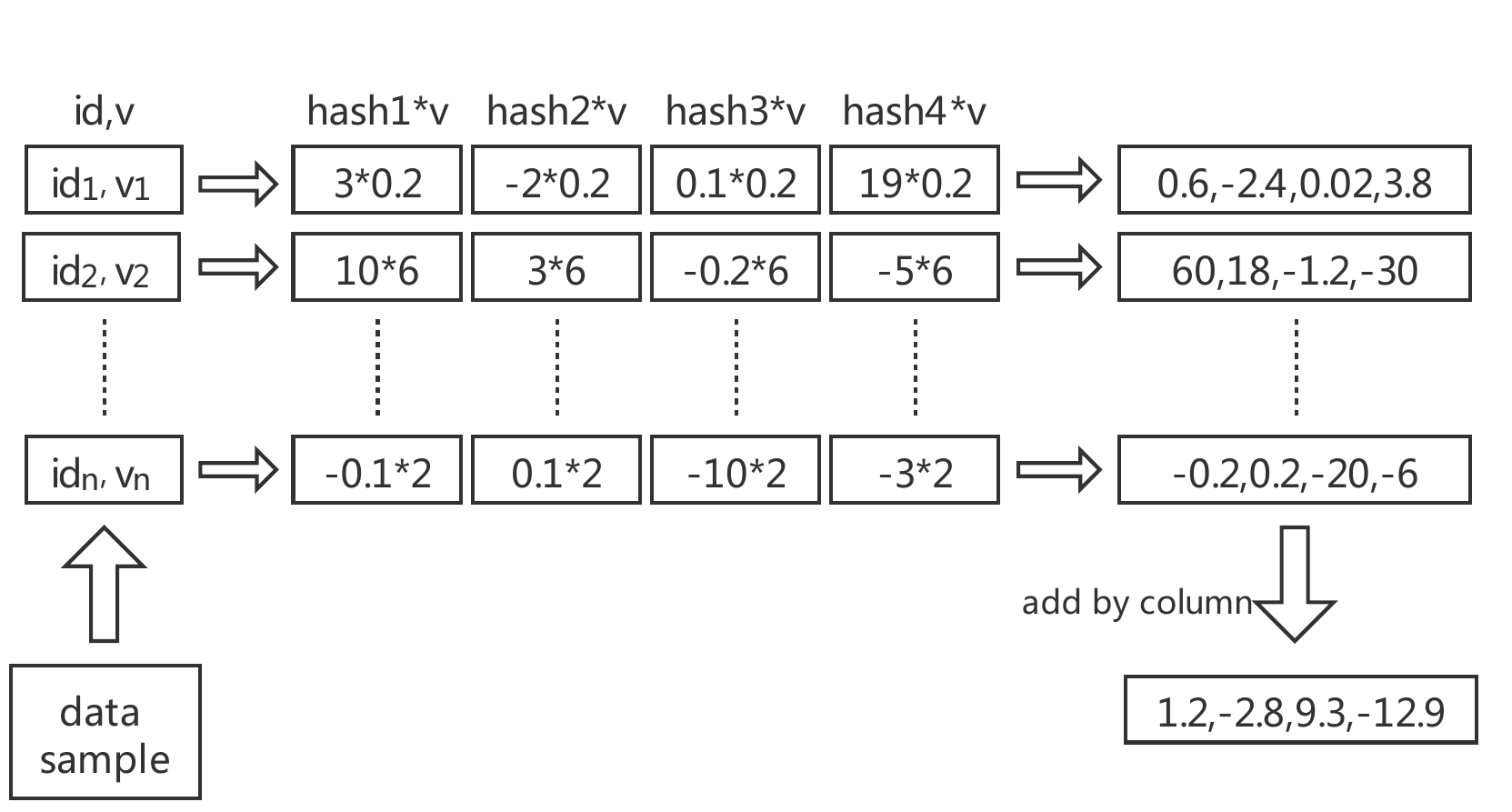}
    \caption{Simhash++ Workflow.}\label{fig:simhash++}
    \vspace{-2mm}
\end{figure}

The number of input neurons of the neural network for high-dimensional data is huge, which will result in severe memory and calculation overhead. To solve this efficiency problem, we extend the Simhash~\cite{sadowski2007simhash} method to Simhash++ that efficiently performs forward calculation for sparse data. Simhash++ generates floating-point weight on-the-fly corresponding to the feature and avoids creating the large neural network. Figure.~\ref{fig:simhash++} illustrates the workflow of Simhash++. There are two changes of Simhash++ over Simhash. Firstly, for a given data sample, Simhash++ hashes each appeared feature name or id $K$ times to get $K$ binary strings by $K$ hash functions, and converts these binary strings to floating-point values obeyed the uniform distribution. Secondly, Simhash++ doesn't binarize the result vector and keep the vector contains floating-point numbers.

\begin{figure}[!h]
    \centering
    \includegraphics[width=1.0\linewidth, trim=50 470 90 40,clip]{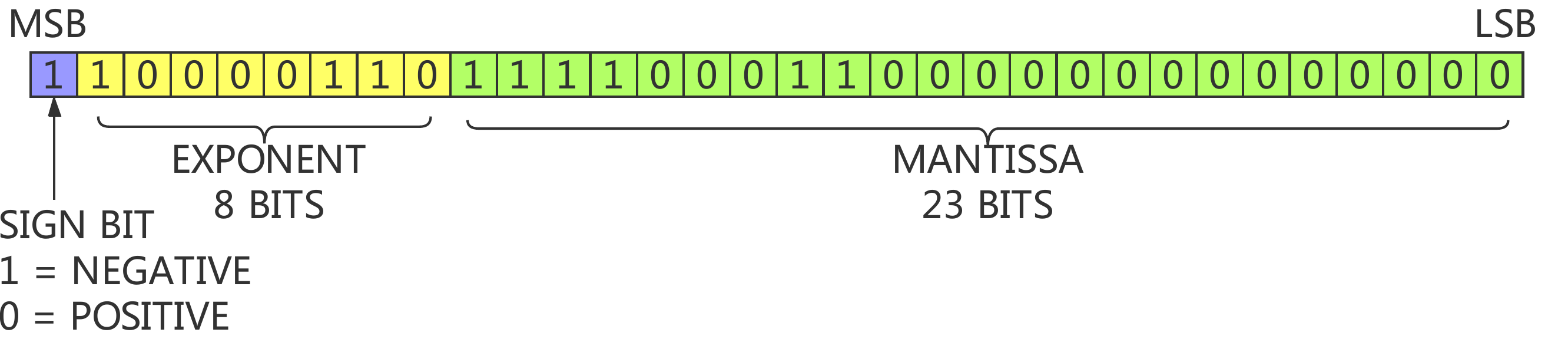}
    \caption{Floating Point Format IEEE 754 for 32 Bits.}\label{fig:ieee754}
    \vspace{-2mm}
\end{figure}

The key magic of Simhash++ is converting a binary string to a floating-point number according to the IEEE 754 standard. Figure.~\ref{fig:ieee754} is an IEEE 754 32-bit floating-point format. From left, the 1st bit is the sign; the 2nd to 9th bits denote the exponent from -127 to +128 and 01111111 denotes the 0 exponent; the remaining 23 bits encode the mantissa in the range of $[1.0,2.0)$. As a result, we use the following two masks to fulfill this conversion:

\vspace{-5mm}
\begin{equation*}
\begin{split}
    Mask1 = 00111111111111111111111111111111 \\
    Mask2 = 00111111100000000000000000000000 \\
\end{split}
\end{equation*}

We perform a bitwise AND with the 32-bit binary string to $Mask1$ to ensure that the sign is positive and then execute an OR operation to $Mask2$ to set the exponent to be 0. Afterward, we get a floating-point number between $[1.0,2.0)$. If the hash functions distribute hash values evenly across the available range, the converted number will obey the uniform distribution, and we can transform the distribution to other ranges such as $[-1.0, 1.0)$. Moreover, the speed of the basic hash functions is critical to the Simhash++. We adopt the xxHash~\cite{xxhash} that is the fastest non-encrypted Hash function as far as we know.

\section{Experiments}\label{sec:experiment}
In this section, we validate the prediction accuracy of GBUN on 8 binary and multi-classification datasets by Python implementation of GBUN. Furthermore, we test the scaling property of GBUN by Spark GBUN on high-dimensional datasets. In these experiments, we take the Xgboost and LightGBM as the benchmark algorithms. We also provide more details about the analysis of experiments and the random seed settings in the section ``More Details about Experiments'' of the technical appendix.

\subsection{Python and Spark Implementations}
Our Python implementation of GBUN depends on PyTorch 2.0+CUDA 10.0 and supports Python 3.X interface. We also implement GBUN based on the Spark 2.4.5 and Rabit~\cite{rabit} distributed communication framework. We adopt many engineering optimization tricks in Python and Spark GBUN to pursue extreme efficiency. However, there is no space to describe these details. 

\subsection{Accuracy Tests}
\begin{table*}[!t]
    \centering
    \caption{Datasets Description and the Evaluation Results in ccuracy Test}\label{tb:data}
    \begin{tabular}{ |l|c|c|c|c||c|c|c|c|c|c|c|}
     \hline
      \multirow{2}{*}{Data} &
      \multirow{2}{*}{\#\textit{Sample}} & 
      \multirow{2}{*}{\#\textit{Feature}} &
      \multirow{2}{*}{Dense} &
      \multirow{2}{*}{\#\textit{Class}} &
      \multicolumn{4}{c|}{Accuracy Metric} &
      \multicolumn{3}{c|}{ Time(second)/Round}
      \\
      \cline{6-12} 
      & & & & & Type & XGB & LGB & GBUN & XGB & LGB & GBUN \\
      
     \hline
     Allstate & 13M:4.3M & 4248 & False & 2 & AUC 
     &.6015& .6032&\textbf{.6036} & 21.86&\textbf{2.57}&36.72\\ 
     \hline
     Epsilon & 400K:100K & 2000 & True &  2 & AUC 
     & .9405&.9451& \textbf{.9596}& 9.01&\textbf{0.54}&0.84\\
     \hline
     KDD12 & 119M:29M & 54M & False & 2 & AUC
     & .7361& .7246&\textbf{.7480}& 218.44& \textbf{159.78}& 247.56\\
     \hline
     News20.b & 139975:999 & 1.3M & False & 2 & AUC 
     & .9885& .9916& \textbf{.9938}& 1.01& \textbf{0.55}& 0.79\\
     \hline
     Rcv1.b & 20243:677K & 47236& False & 2 & AUC 
     & .9921& \textbf{.9937}& .9932& \textbf{0.10}& 0.12& 2.50\\
     \hline
     \hline
     Aloi & 75600:32400 & 128 & True & 1000 & mAP 
     & .9507& .9084& \textbf{.9651} & 20.36& 5.50& \textbf{2.73}\\
     \hline
     Letter & 10500:5000 & 16 & False & 26 & mAP 
     & .9517& .9615& \textbf{.9618}& 0.04& \textbf{0.03}& 0.55\\
     \hline
     News20.m & 62061:15935 & 3072 & False & 20 & mAP 
     & .7919& .8092& \textbf{.8505}& 1.01& 1.22& \textbf{0.59}\\
     \hline
    \end{tabular}
\end{table*}



The details of 8 data sets for the accuracy test are listed in Table.~\ref{tb:data}. All the datasets are selected from Libsvm Datasets~\cite{LibsvmData} except the Allstate~\cite{allstate}. Allstate, Epsilon~\cite{epsilon}, KDD2012~\cite{KDD12}, News20.binary~\cite{lang1995news20multi}, and Rcv1.binary~\cite{lewis2004rcv1} datasets are employed to test the prediction accuracy on the binary classification. These datasets are relatively large in scale. Some have a large number of samples, such as Allstate; some have a large number of features, such as News20.binary and Rcv1.binary; and some have both, such as KDD2012. Besides, espilon is the dense dataset with the largest number of features in Libsvm Datasets. The remaining 3 datasets Aloi~\cite{rocha2014aloi}, Letter~\cite{letter} and News20.multi~\cite{lang1995news20multi} are all multi-class datasets, and the numbers of classes are all over 20. There is another Rcv1.m multi-class dataset with 53 classes in Libsvm Datasets, but two classes in the test set did not appear in the training set. None of the three algorithms can handle this dataset, so we exclude it in the experiment. For the binary classification tasks, we use AUC as the evaluation metric, and for the multi-class classification tasks, we use mAP (Mean Average Precision) as the evaluation metric.


Our accuracy test environment is a machine with 24 cores(Intel(R) Xeon(R) CPU E5-2683 v4 @ 2.10GHz),  128G memory and a P40 graphics card. We test the three algorithms GBUN, Xgboost, and LightGBM on the above 8 datasets, and all experiments are enabled GPU acceleration. Both Xgboost and LightGBM(EFB disabled) use the exact greedy algorithm for searching split points. We perform the grid search strategy to tune three parameters, model size, shrinkage rate, and L2 regularization weight, for all the three algorithms. The model size parameters of GBUN, Xgboost, and LightGBM are the number of output neurons of the untrained neural network($K$), the deepest tree depth, and the maximum number of leaves, respectively. The possible parameter values of the model size of both GBUN and LightGBM are $2^5, 2^6, 2^7, 2^8, 2^ 9,2^{10}$, and also set the maximum tree depth to be infinite for LightGBM. The range of the deepest tree depth of Xgboost is $[5,10]$.  The shrinkage and L2 regularization weight of the three algorithms range from 0 to 1 with 0.1 interval. However, the L2 regularization weight is fixed at 0 for datasets with more than 1 million training instances to reduce the parameter tuning cost. The number of boosting in the training process is 300, and we take the best test AUC in the 300 rounds for each evaluation as the final accuracy result. Besides, tuning parameters on large data sets are very time-consuming, so we exclude the parameter combinations which train more than 24 hours.

\begin{figure}[!p]
    \centering
    \subfloat[Allstate]{
        \includegraphics[width=0.5\linewidth, trim=10 20 50 45,clip]{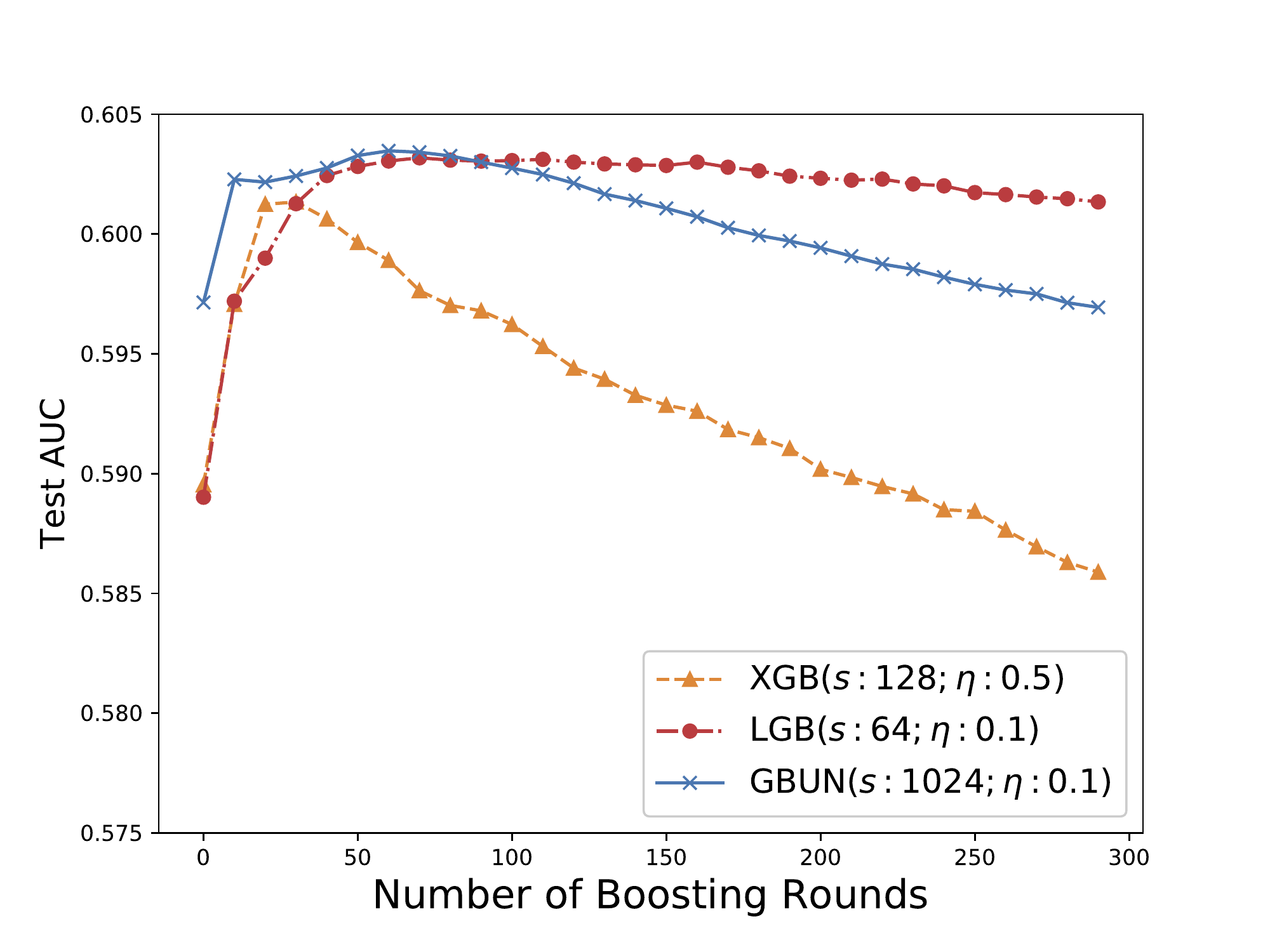}
        \label{fig:exp-allstate} }
    \subfloat[Epsilon]{
        \includegraphics[width=0.5\linewidth, trim=10 20 50 45,clip]{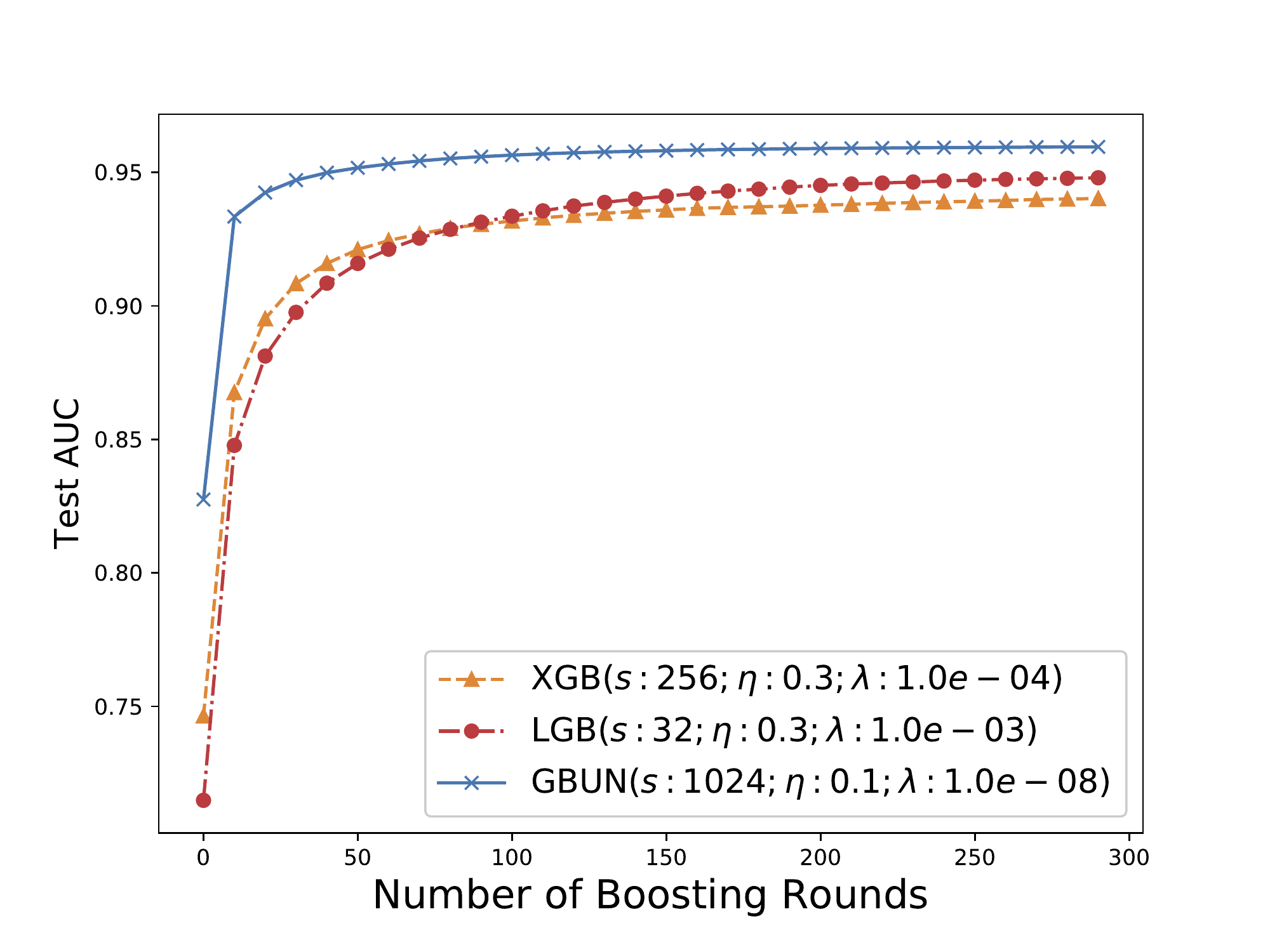}
        \label{fig:exp-epsilon} }
    \vspace{-3mm}
    \subfloat[KDD12]{
        \includegraphics[width=0.5\linewidth, trim=10 20 50 45,clip]{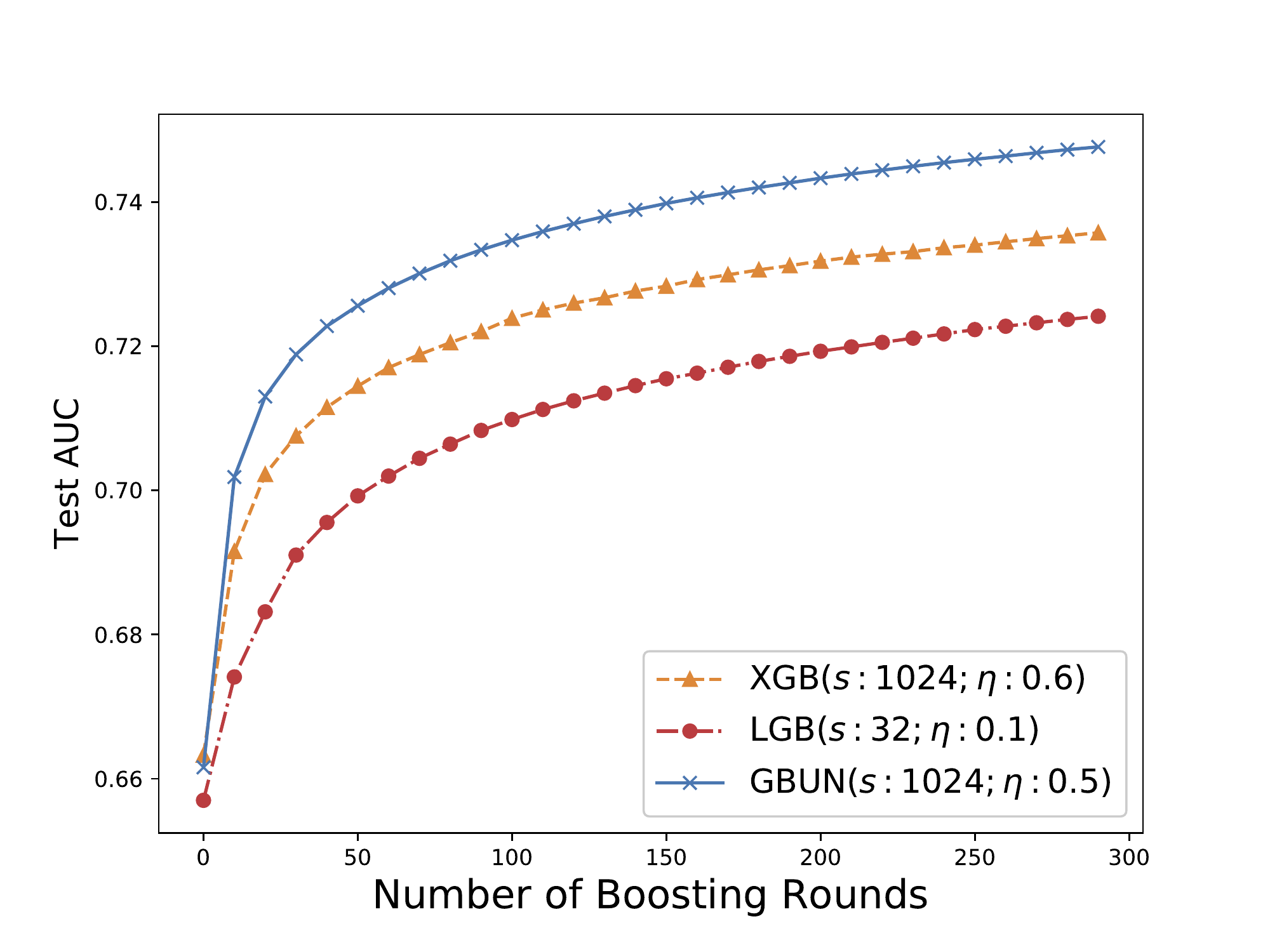}
        \label{fig:exp-kdd12} }
    \subfloat[News20.b]{
        \includegraphics[width=0.5\linewidth, trim=10 20 50 45,clip]{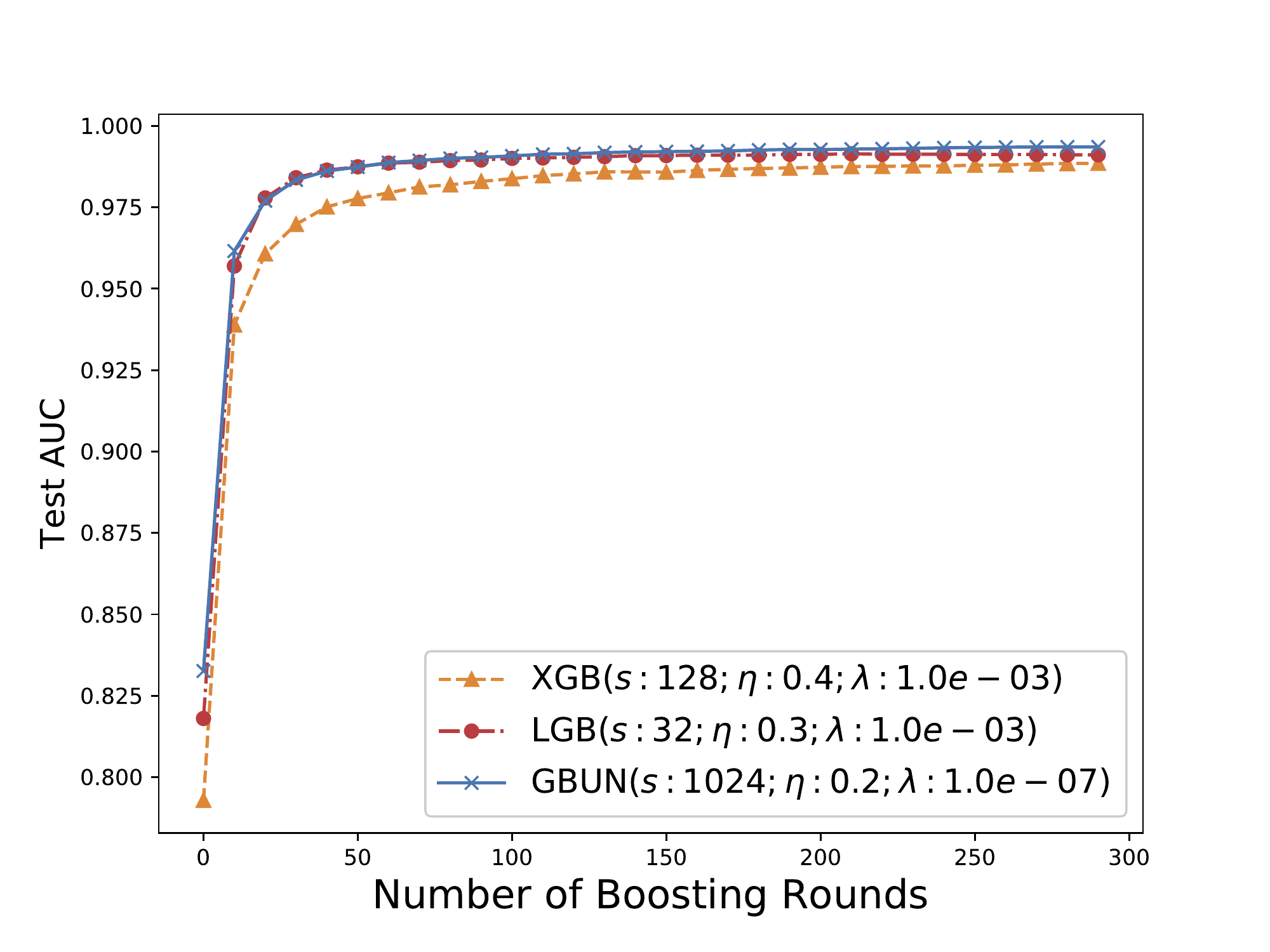}
        \label{fig:exp-news20.b} }
    \vspace{-3mm}
    \subfloat[Rcv1.b]{
        \includegraphics[width=0.5\linewidth, trim=10 20 50 45,clip]{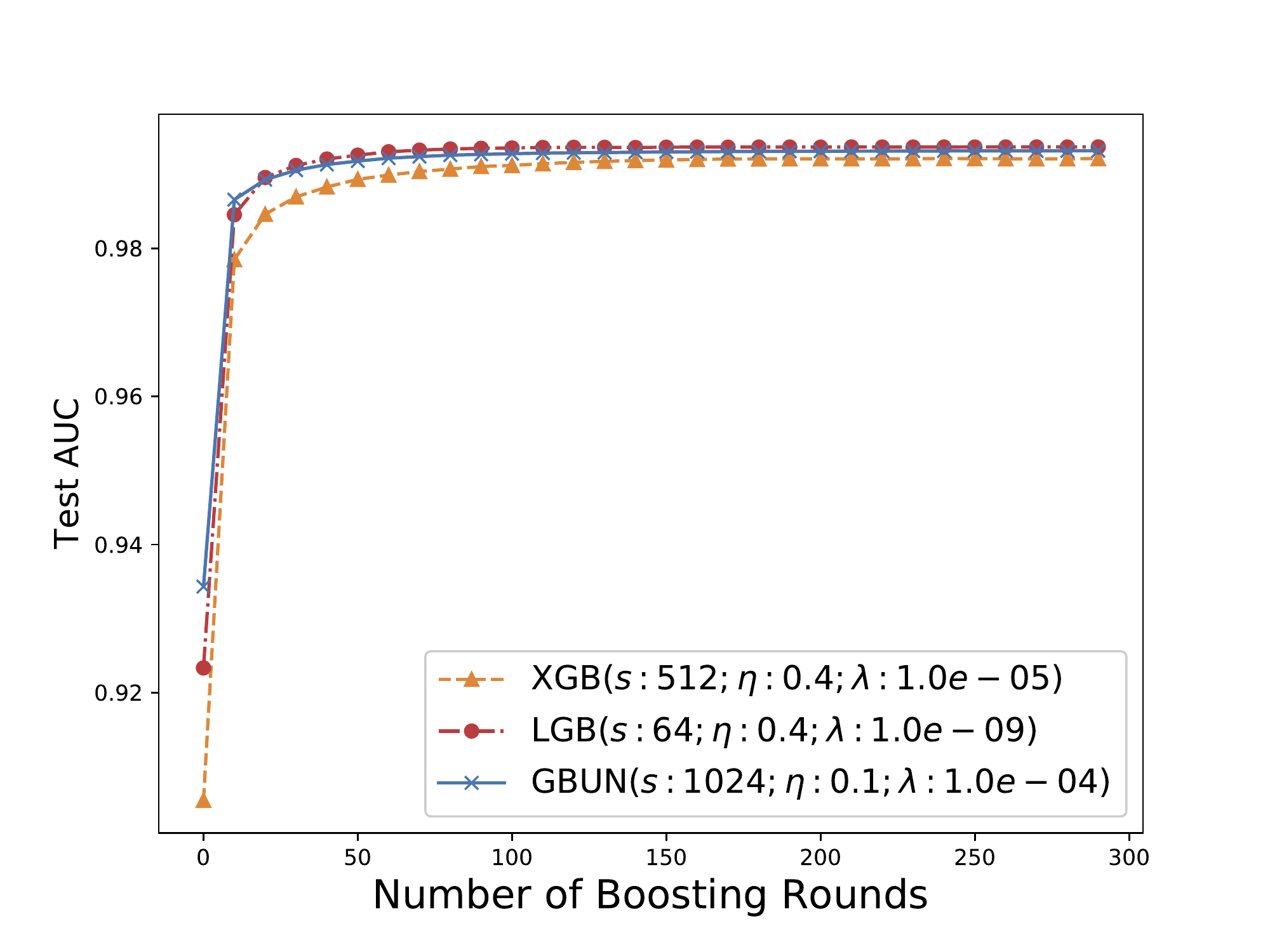}
        \label{fig:exp-rcv1} }
    \subfloat[Aloi]{
        \includegraphics[width=0.5\linewidth, trim=10 20 50 45,clip]{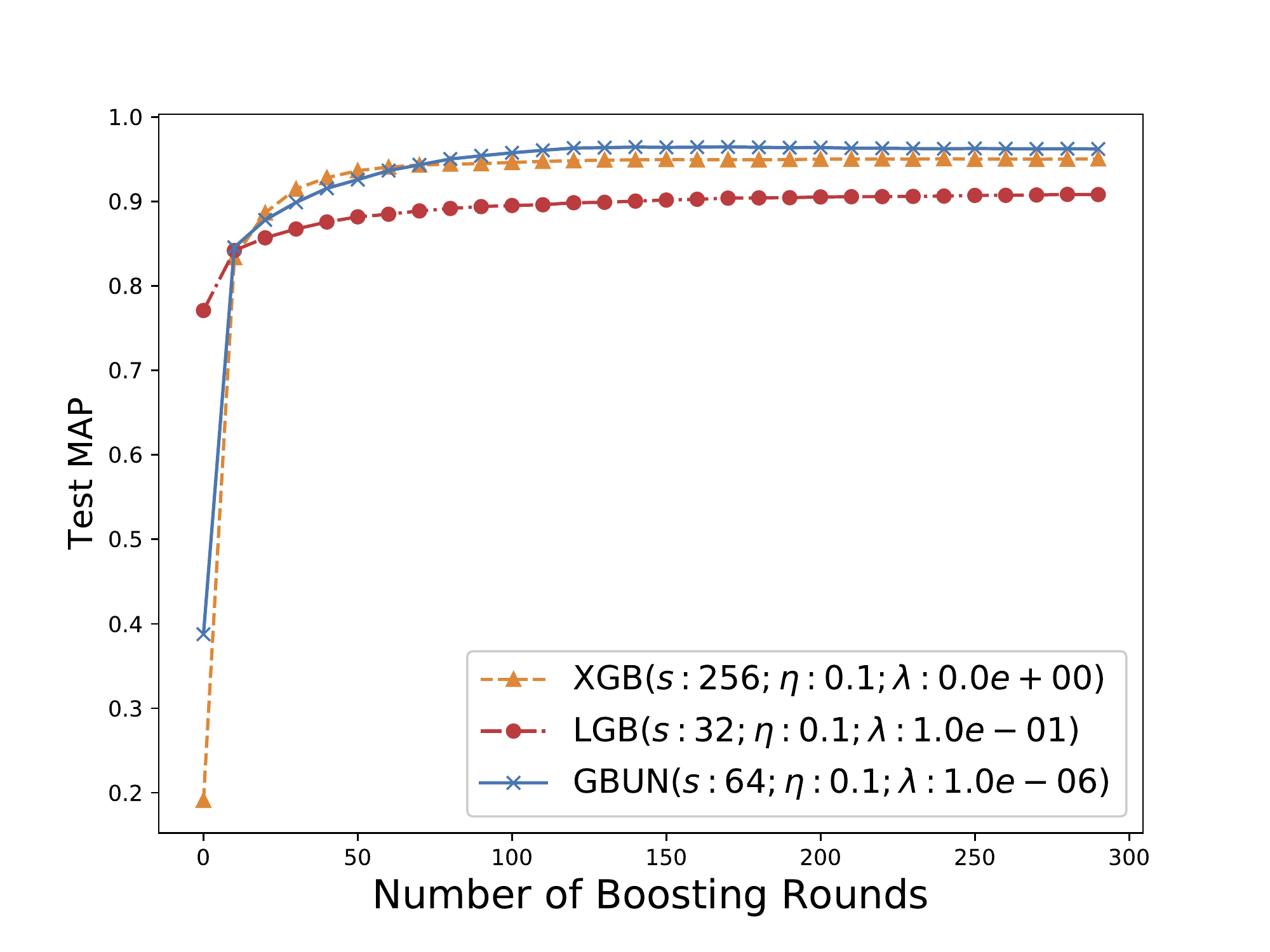}
        \label{fig:exp-aloi} }
    \vspace{-3mm}
    \subfloat[Letter]{
        \includegraphics[width=0.5\linewidth, trim=10 20 50 45,clip]{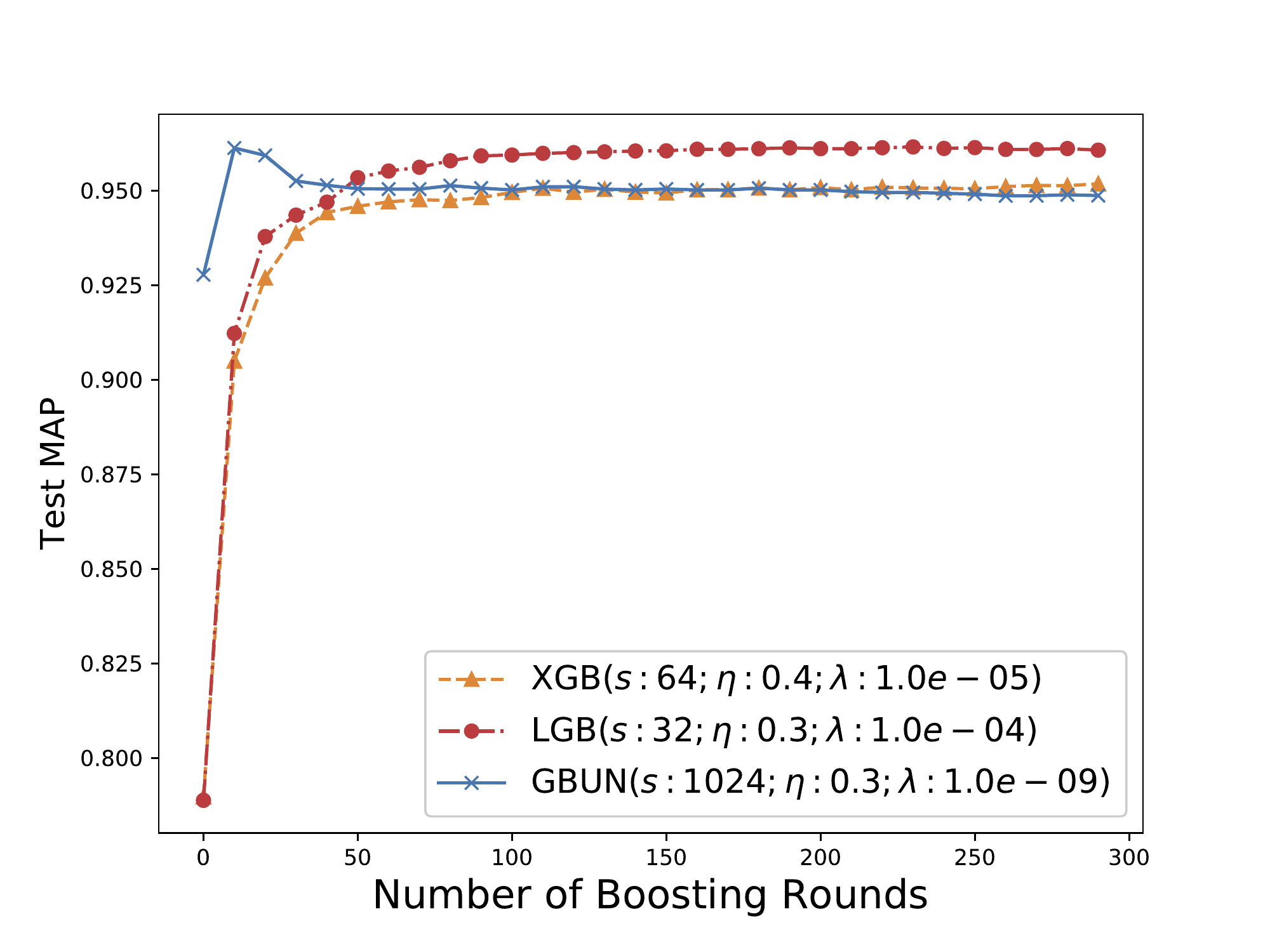}
        \label{fig:Letter} }
    \subfloat[News20.m]{
        \includegraphics[width=0.5\linewidth, trim=10 20 50 45,clip]{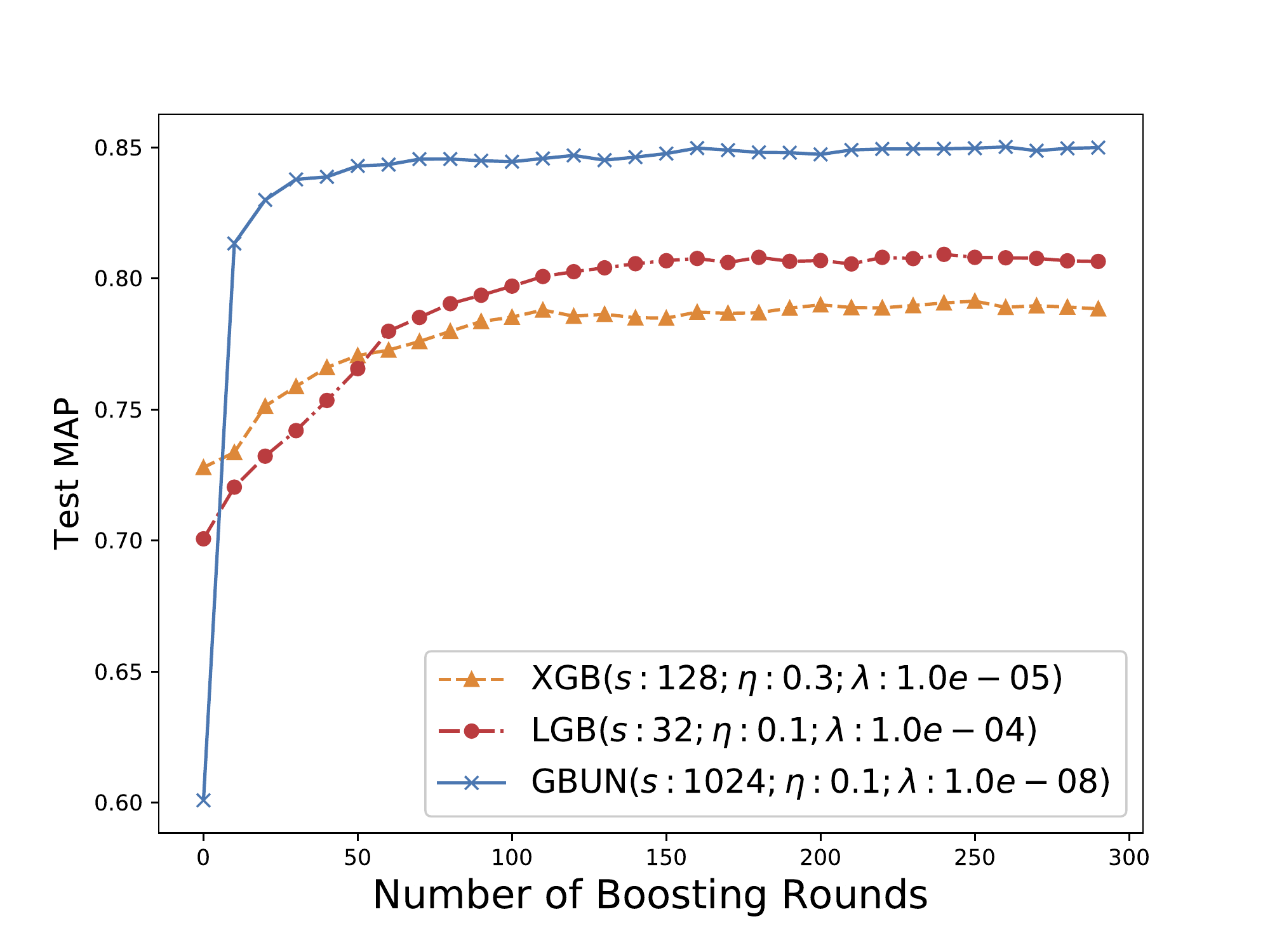}
        \label{fig:exp-news20.m} }
    \vspace{-3mm}
    \caption{Convergence curves of the three algorithms with the optimal parameters on each dataset. Here $s$ is the model size parameter, $\eta$ is the shrinkage rate, and $\lambda$ is the L2 regularization parameter. Since the $\lambda$s for the training datasets with more than 1 million samples are fixed to 0, we omit them in these figures.
    }\label{fig:exp-curves}
\end{figure}

We provide the accuracy test results in Table.~\ref{tb:data} and Figure.~\ref{fig:exp-curves}. In the table, we also record the average per-round time costs of these tests. The Figure.~\ref{fig:exp-curves} shows the convergence curves of the accuracy metric on the test set with the optimal accuracy and the corresponding parameters of each experiment(for datasets with more than 1 million training samples, we fix the L2 regularization parameter to be 0). From the table, we can see that GBUN achieved the best accuracy on 7 datasets except for Rcv1.b. The figure also illustrates that the convergence rate is very competitive compared to Xgboost and LightGBM. On the whole, GBUN is at least as good as Xgboost and LightGBM in the accuracy. In addition, we find out that to achieve the best accuracy result, GBUN needs a bigger model size compared to Xgboost and LightGBM, which may raise more training time costs than the other two algorithms.



\subsection{Distributed Experiments}
The distributed experiment environment is Amazon EC2 Spot and the Spark version is 2.4.5. Except for testing algorithms' performance with different model sizes, the three algorithms all use default parameters. We use the average training time of 100 rounds to measure training performance. We only test the scaling property of GBUN in the distributed experiments, because the distributed implementation of GBUN is equivalent to the basic GBUN in mathematics.

\begin{figure}[!]
    \centering
    \subfloat[Model Size 64]{
        \includegraphics[width=0.5\linewidth, trim=10 0 10 0,clip]{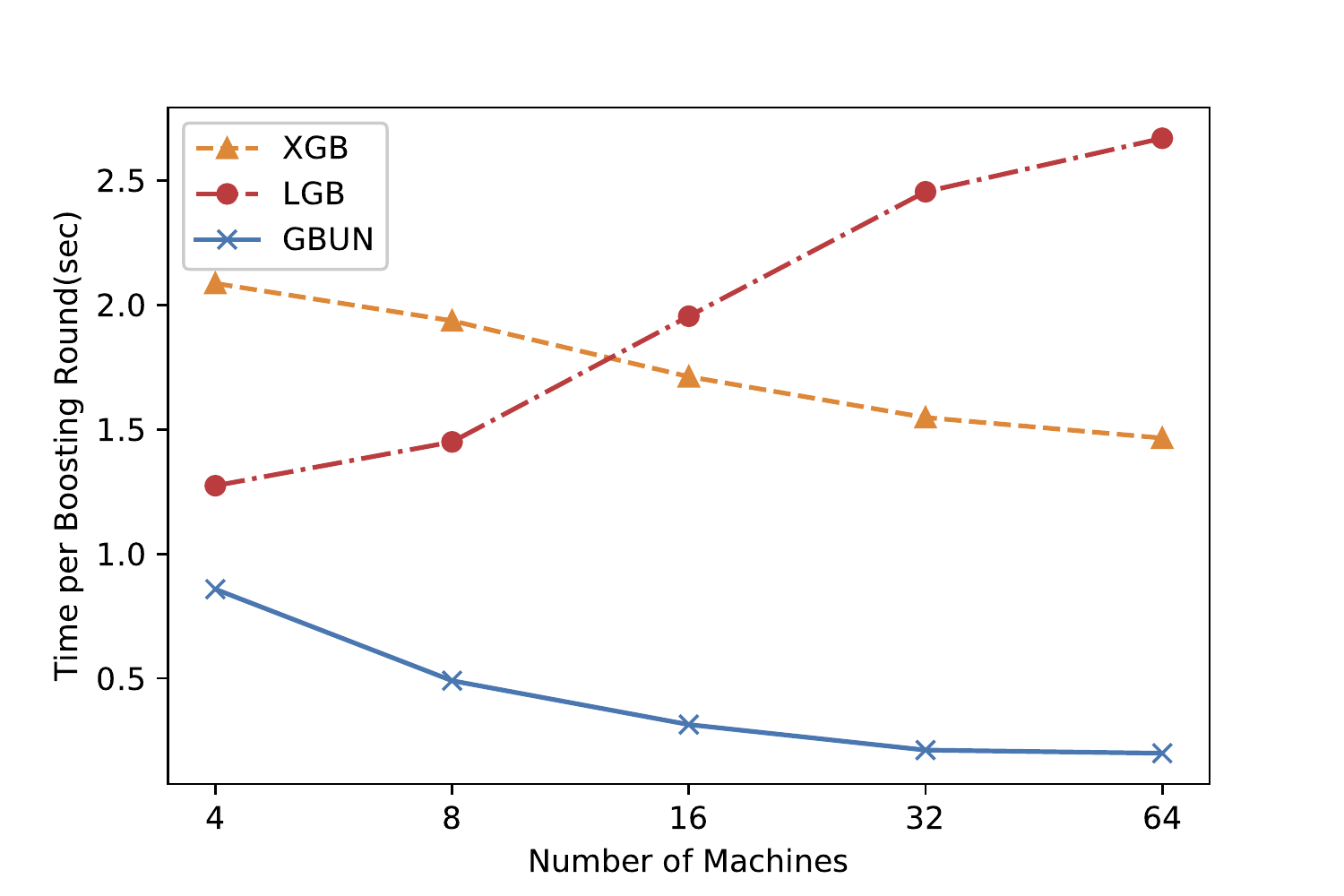}
        }
    \subfloat[Model Size 1024]{
        \includegraphics[width=0.5\linewidth, trim=10 0 10 0, clip]{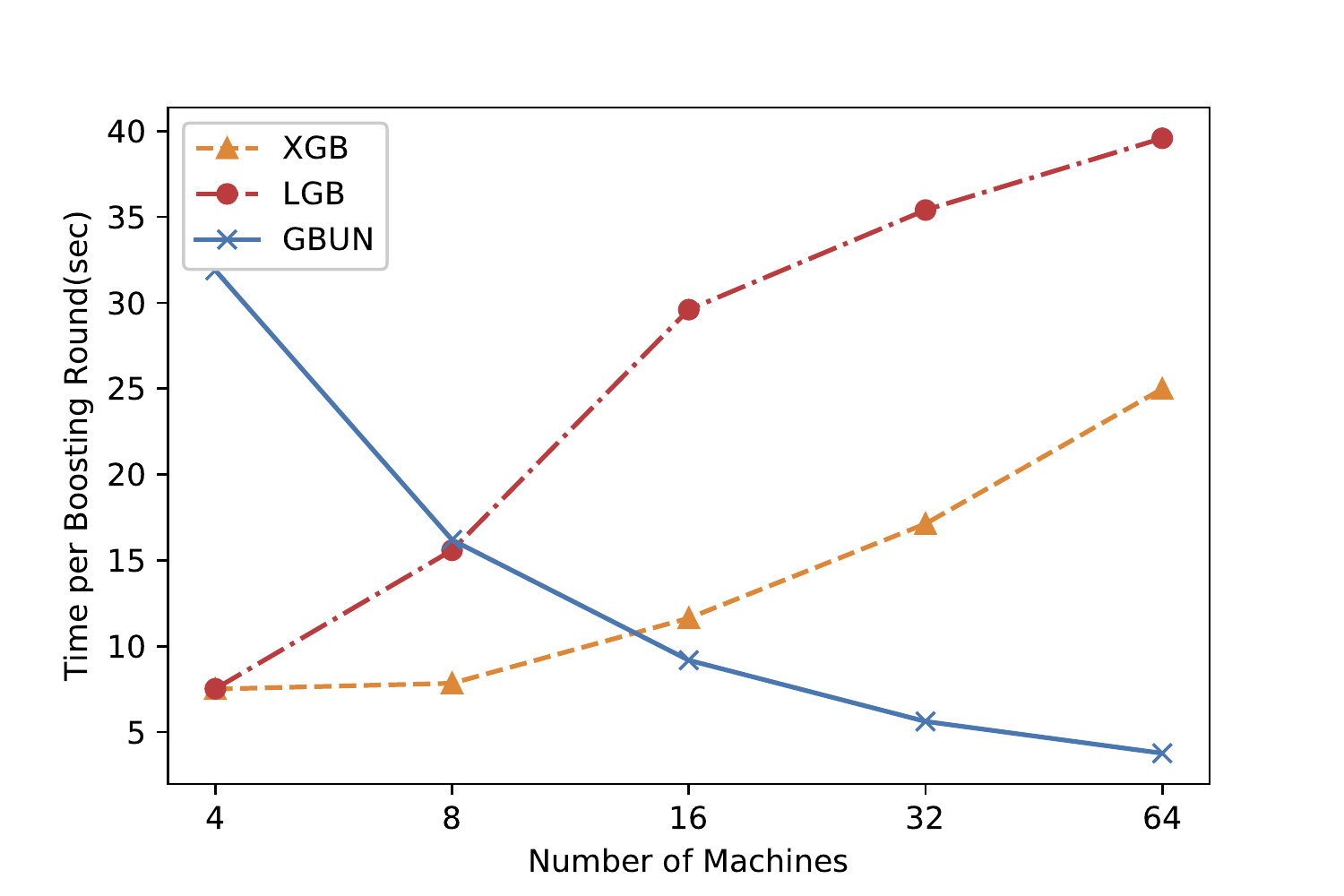}
        }
    \vspace{-3mm}
    \caption{Training time costs of the three algorithms on the KDD12.sub dataset over clusters with different sizes.}\label{fig:dist-kdd12-sub-curvers}
    
\end{figure}

We should test the scaling property of GBUN comparing to Xgboost and LightGBM. However, In our experiment environment, Spark Xgboost can only handle datasets with hundreds of thousands of features at most, while Spark LightGBM can process data sets with millions of features because of the EFB technique. Therefore we need a dataset with feature dimensions in the order of hundreds of thousands. Since there is no suitable dataset in Libsvm Datasets, we sample 1\% of the KDD12 training dataset and then copy it 10 times. So we get a dataset with about 390K features and 1.2 million samples and named the dataset as KDD12.sub. On this dataset, we test the training time cost of the three algorithms on clusters of 4, 8, 16, 32, and 64 machines with 2CPU and 8GB memory. The number of Spark executors is 1 for Xgboost(officially recommended setting for multi-core acceleration), and 2 for both LightGBM and GBUN. Figure.~\ref{fig:dist-kdd12-sub-curvers} shows how the average per-round time costs of three algorithms changes under two model sizes (64 and 1024). We can see that the training time costs of GBUN on both model sizes decline steadily and much faster than Xgboost and LightGBM. The training time costs of Xgboost(64 model size) and LightGBM(64 and 1024 model size) rise rapidly with machines. Comparing with Xgboost on the clusters with 4, 8, 16, 32, and 64 machines, GBUN speeds up to 2.4x, 3.9x, 5.4x, 7.3x, and 7.3x when the model size is 64; It speeds up to 0.2x, 0.5x, 1.3x, 3.0x, and 6.6x while the model size is 1024. By comparing with LightGBM, GBUN speeds up to 1.5x, 3.0x, 6.2x, 11.6x, 13.4x when the model size is 64; It speeds up to 1.0x, 3.2x, 6.3x, and 10.5x while the model size is 1024. Obviously, the scaling property of Spark GBUN has an overwhelming advantage over Xgboost and LightGBM. 

\begin{figure}[!]
    \centering
    \includegraphics[width=1\linewidth]{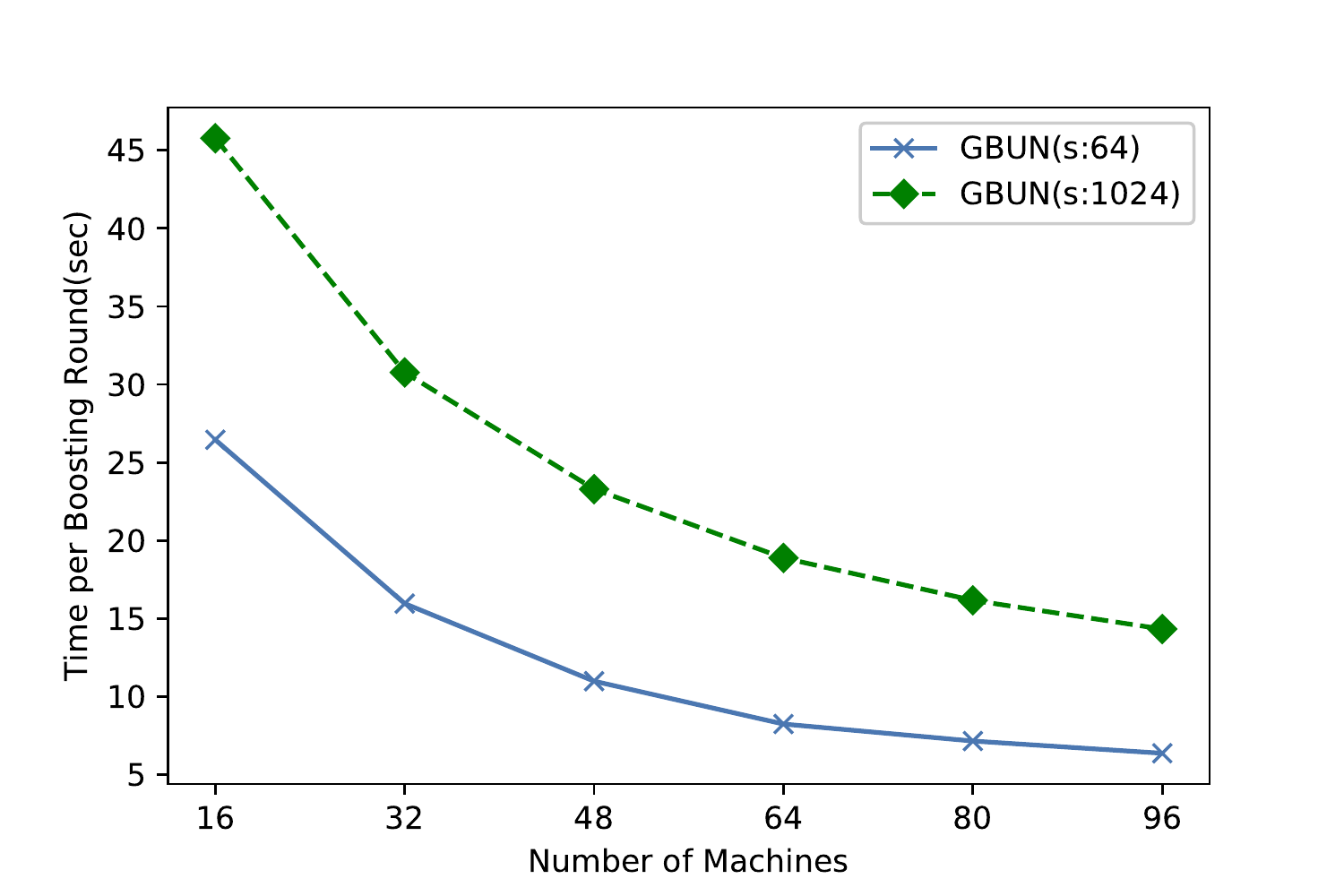}
    \caption{Training time costs of GBUN on the KDD12 dataset over clusters with different sizes.}\label{fig:exp-dist-kdd12}
    \vspace{-2mm}
\end{figure}

\begin{figure}[!]
    \centering
    \subfloat[Model Size 64]{
        \includegraphics[width=0.5\linewidth, trim=10 0 10 0,clip]{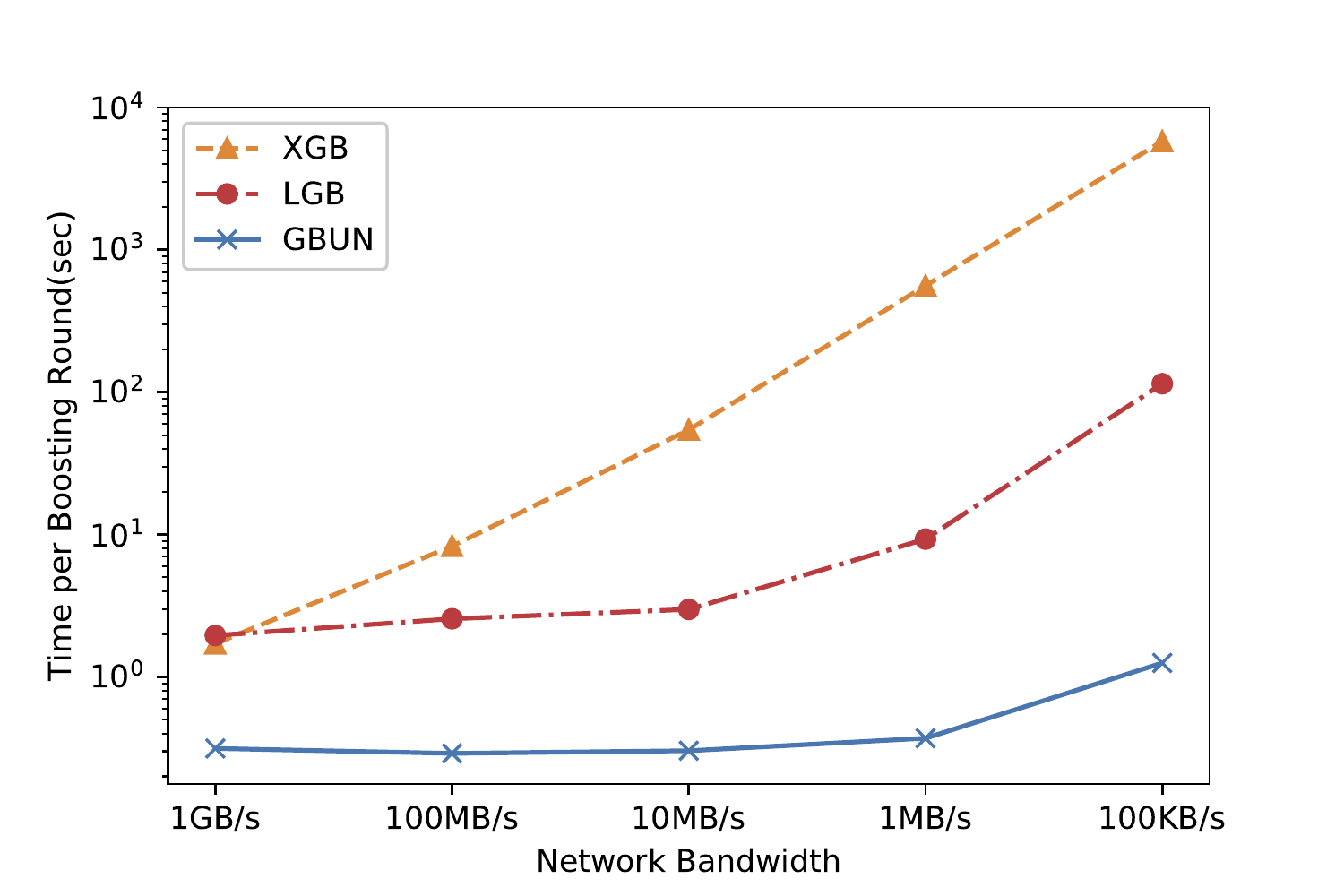}
    }
    \subfloat[Model Size 1024]{
        \includegraphics[width=0.5\linewidth, trim=10 0 10 0, clip]{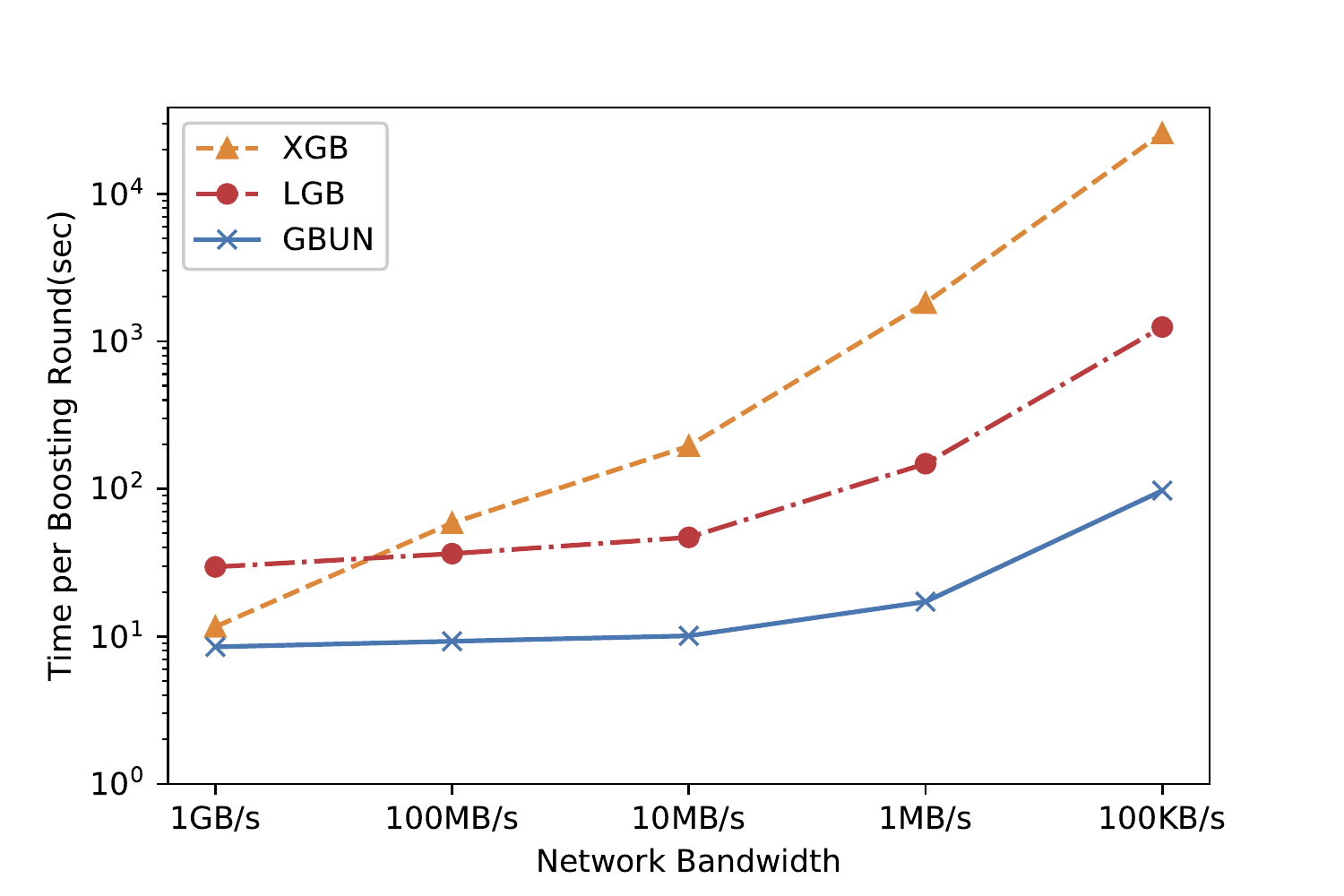}
        }
    \vspace{-3mm}
    \caption{Training time costs of the three algorithms on the KDD12.sub dataset over the 16 machine cluster with different network bandwidths.}\label{fig:netlimit}
\end{figure}

We also test the scaling property of GBUN on the ultra-high dimensional dataset KDD12 that can't be handled by Spark Xgboost and LightGBM. On the KDD12 dataset, we check the training time of Spark GBUN on clusters with 16, 32, 48, 64, 80, and 96 machines. The machine configurations of 16 and 32 machine clusters are 48C196G and 32C128G. The machine configuration for the clusters with 48 to 96 machines is 16C64G. We use 48C and 32C machines for small size clusters because there are no 16C192G and 16C128G instances on Amazon Spot EC2 to ensure that each machine on the small clusters can hold its data partition in memory. However, Spark GBUN does not enable any multi-core acceleration capabilities. We just set the number of Spark executors to 16 for all clusters to ensure each machine's available computing power in all the tests are the same. Figure.~\ref{fig:exp-dist-kdd12} illustrates the average per-round training time costs also decrease steadily on both 64 and 1024 model sizes. On the ultra-high dimensional dataset, the scaling property of GBUN is still decent.

In order to validate the performance of distributed GBUN under the low network bandwidth environment, by WonderShaper 1.4.1~\cite{WonderShaper} we limit the bandwidth of the cluster with 16 machines(2C8G) to 1GB/s, 100MB/s, 10MB/s, 10MB/s, 1MB/s, and 100KB/s. We validate how these different bandwidths affect the performance of the three distributed algorithms. Figure.~\ref{fig:netlimit} shows each algorithm's training time under different network bandwidth limits on the KDD12.sub dataset. We can find out that the time increase on GBUN is much less than that on Xgboost and LightGBM. Comparing with Xgboost on the cluster with 1GB/s, 100MB/s, 10MB/s, 1MB/s, and 100KB/s bandwidth, GBUN speeds up to 5.4x, 28.6x, 179.7x, 1510.5x, and 4613.8x when the model size is 64; it speeds up to 0.2x, 0.5x, 1.3x, 3.0x, and 6.6x while the model size is 1024. By comparing with LightGBM, GBUN speeds up to 1.3x, 6.3x, 19.3x, 105.9x, and 264.4x when the model size is 64; it speeds up to 3.2x, 3.9x, 4.6x, 8.6x, and 12.8x while the model size is 1024 respectively. Apparently, GBUN is much more efficient than Xgboost and LightGBM in the distributed environment with low network bandwidth and can be employed in the federated learning applications among many mobile and edge devices connected by public internet.

\section{Conclusion}
This paper has proposed a novel gradient boosting algorithm called GBUN, which uses the untrained randomly generated neural network as the base predictor to reduce much communication costs for distributed learning. Enabling GBUN to handle high-dimensional sparse data, we have extended Simhash algorithm to mimic forward calculation of the neural network. The most significant advantage of GBUN is that the communication overhead for distributed learning is tiny. 

In the future, we will study how the structure and weights of the untrained neural network affect the prediction accuracy in theory, then adopt untrained randomly generated deep neural networks and better initialization methods to improve the accuracy of GBUN. Furthermore, we will try to turn GBUN into a high-performance federated learning algorithm for the scenarios with a large number of mobile and edge devices connected via the public mobile internet.

\begin{appendices}
In this technical appendix, we provide supplement description about normalizing the untrained neural network's outputs, initializing untrained neural network, validating the uniformity of the floating-point numbers generated by our hash trick, the engineering tricks of our implementations, and more details about the experimental results.

\section{Normalization and Initialization} 
The purpose of the general neural network algorithm's normalization~\cite{ioffe2015batch} and initialization\cite{glorot2010init,he2015delving} is to eliminate the gradient vanishing and explosion in the backward propagation by controlling the output distribution of the neural network. GBUN doesn't train the neural network, so the purpose of normalization and initialization in GBUN is different from the general neural network. However, as a new algorithm, there are still many unknowns about GBUN and no theory to guide the normalization and initialization. At present, our works about normalization and initialization are based on intuition and empirical study.

The normalization method currently used by GBUN is straightforward. Equation.(1) in the main paper indicates that the normalization makes the mean and standard deviation of the distribution of the output $\tilde{\mathbf{z}}_{\cdot j} $ be 0 and 1 respectively. The intuitive goal of the normalization ensures that the distribution of the multivariate probability distribution of the network outputs activated by $softmax$ is as uniform as possible. If the distribution is not uniform enough, it also means that the projections of the training samples by the neural network and the activation function concentrates in some small parts of the possible probability space, which reduces the discrimination between the samples and is harmful to the subsequent gradient boosting.

As we all know, The distribution of the multivariate probability distribution is the Dirichlet distribution. The uniform Dirichlet distribution requires all the concentration parameters to be 1, which is our intuitive normalization goal. Based on some empirical tests, we find out that the outputs normalized by Equation.(1) are approaching the goal. Besides, since the normalization always turns the mean of the outputs to 0, we don't introduce the bias into the untrained neural network. 

Compared to normalization, the initialization goal is not explicit by now. We intuitively think that the weight matrix of the neural network should be as different as possible. But such as: How to define the similarity among the matrices? How to guide weight matrix initialization based on the possible similarity measurement? We have no answers to these problems at present. However, we empirically test several common random distributions, such as the uniform distribution between -1 and 1, the standard normal distribution, and the mixed Gaussian distribution. The results show that the network generated by a uniform distribution between -1 and 1 achieves the best accuracy overall. When the dataset is dense, we can get more accurate result by randomly setting 90\% of the weights in the matrix as 0. Therefore, in the Python and Spark implementations, we used the above neural network initialization method.

\begin{figure}[!]
    \centering
    \includegraphics[width=0.8\linewidth]{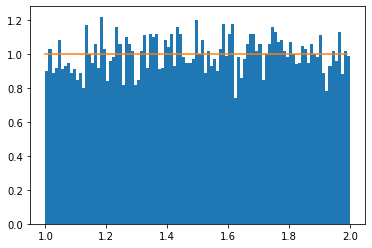}
    \caption{After hashing the integers from 1 to 10000, we convert the hash values to floating-point numbers that obeys a uniform distribution, and then draw a histogram of 100 bins.}\label{fig:distribution}
    \vspace{-2mm}
\end{figure}

\section{Uniformity of Generated Floating-point Values}\label{sec:sparse}

In the section ``Forward High-dimensional Sparse Data'' of the main paper, we introduce a method to generate floating-point values obeying the uniform distribution based on the binary hash codes. We generate 10,000 32-bit floating-point numbers by this method to validate their uniformity. Figure.~\ref{fig:distribution} shows that the generated 10,000 floating-point numbers are distributed uniformly in the range $[1.0, 2.0)$. It is in line with our expectation.

\section{Implementations}\label{sec:python}
We implement both Python and Spark GBUN libraries. The Python library is an efficient stand-alone machine learning tool with GPU acceleration, and the Spark GBUN targets to tackle large-scale datasets on large computing clusters. 

\subsection{Python Library}
Because there are many matrix operations in the GBUN algorithm, we implement the Python GBUN based on Pytorch that supports tensor calculations, linear algebra operations, and GPU acceleration. Besides, based on the Autograd mechanism of Pytorch, users can customize objective functions without deriving the gradient formulas. According to the Algorithm.~1, the basic GBUN algorithm can be implemented based on Pytorch with a few lines of code. However, there are some limitations to the naive implementation.

First of all, the simple GBUN implementation cannot support GPU acceleration for big data due to the limited GPU memory capacity. Based on the GBUN distributed learning principle, we implement the batch training mode with GPU acceleration. In each boosting, batch training method handles the training dataset on GPU batch by batch. The mathematical principle of batch training method is the same to the distributed learning approach of GBUN. However, repeatedly moving data from main memory to GPU memory raises much time costs. Our analysis shows that the main overhead is caused by repeatedly allocating and releasing memory structures on the GPU during. We introduce a memory pre-allocating mechanism to decrease this overhead: allocate memory structures for data batches and the intermediate data before training and reuse them. Consequently, the batch training GBUN removes the overhead of repeatedly allocating and releasing memory structures and improves the batch learning mode's efficiency. 

Secondly, the naive implementation cannot efficiently handle sparse data. The matrix multiplication is the core operation of the neural network forwarding. However, the sparse matrix multiplication is inefficient in Pytorch. Usually, only if the matrix's sparsity ratio is more than 98.5\%, the sparse matrix multiplication executes faster than the dense matrix multiplication on the same scale~\cite{SparseMatrixInPyTorch}. Therefore, we implement the Simhash++ method proposed in the section Forward High-dimensional Sparse Data to efficiently forward the sparse data. Furthermore, it is difficult to slice the default sparse tensor(COO Tensor) of Pytorch to support batch training mode. We implement the compressed sparse row tensor (CSR tensor) to overcome this problem, which only supports two-dimensional tensor, namely, matrix. The internal structure of the CSR tensor is composed of three one-dimensional tensors, ``data'',``indptr'', and ``indices''. ``data'' holds the non-zero values;``indptr'' stores the row indices; ``indices'' indicates the column indices of non-zero values. The lengths of the ``data'' and ``indices'' are the number of non-zero(nnz) values, and the length of ``indptr'' is the number of rows plus one. Although the CSR tensor is suitable for slicing, there is still a trouble for the memory pre-allocation mechanism. Usually, some batches contain more non-zero values, and some hold less. The row slicing method wastes the GPU memory because we should pre-allocate memory based on the worst case. So, our slicing strategy aims to split the non-zero values evenly. Let $\#nnz\_batch$ be the pre-allocated nnz, $\#nnz$ is the nnz of the whole training data set, and $\#batch$ is the number of batches, $row\_nnz\_max$ is the row with maximum nnz. Then there are,

$$\#nnz\_batch = max(row\_nnz\_max, \frac{\#nnz}{\#batch})$$

We expect the nnz in batches are as uniform as possible, but the pre-allocated memory must at least hold one complete row. The lengths of ``data'' and ``indices'' in the pre-allocated CSR tensor on the memory are equal to $\#nnz _batch$. Also, we should determine the length of "indptr" in the pre-allocated CSR tensor, limiting the maximum number of rows in a batch. Since some rows may contain few non-zero values and some batches may hold more rows than others, the "indptr" should keep more capacity than the batches' average rows. Therefore, with $\#row$ as the total number of rows in the dataset, and $\#row\_nnz\_25q$ representing 25\% quantile of the row nnz, we set the maximum row capacity to:
$$\#indptr-1 = \frac{\#row}{\#row\_nnz\_25q}$$
Although the above setting cannot meet the most extreme cases, it works for most real cases.

Besides, to ensure the training process to be reproducible, we also expose the random seed of matrix generation as a parameter for dense data. In the inner of the Python library, we use       ``torch.manual\_seed(seed)'' function to set the random seed of the Pytorch. By default, we set the random seed to be 0. For sparse data, since we use Simhash++ method to mimic the forward calculation, we use the following rule to set the seeds of xxHash~\cite{xxhash}: For each boosting, we hash the feature id concatenated with the round number, and the $K$ seeds of the hash functions are from 0 to $K-1$. 

The current Python implementation of GBUN depends on PyTorch 2.0+CUDA 10.0, and also supports Python 3.X interface. 

\subsection{Spark Library}\label{sec:spark}
Spark is a widely used big data processing system, and we implement distributed GBUN based on the 2.4.5 version. Both Xgboost and LightGBM have provided Spark library. But the communication bottleneck of the distributed GBDT causes the inefficiency in handling large-scale high-dimensional learning tasks. In comparison, the distributed GBUN produces much smaller communication volume. To fulfil the expected high performance of the distributed GBUN, we optimize the performance of Spark GBUN on 3 aspects.

Firstly, according to the steps of distributed GBUN, it is easy to implement based on the Map-Reduce programming paradigm. However, since the Spark RDD data set is immutable, there are some difficulties in updating the predicting scores of training samples after each boosting. Grouping the features, labels, and predicting scores together should create a new RDD for each boosting. Otherwise, holding the predicting scores exclusive leads to join the predicting score RDD and the training data RDD every boosting. Another method is that we can hold the predicting scores in the driver's memory, but the driver's limited memory also limits the number of training samples. All these approaches are inefficient.

To avoid these problems, Spark GBUN adopts the same programming model as Xgboost and LightGBM: the whole training procedure executes in a single ``mapPartitions'' transformation, and the number of executors must be equals to the number of data partitions to ensure training all the partitions on the same time. Actually, in this way, Spark only distributes the data partitions to computing nodes, and the communication among nodes depends on other communication frameworks. Xgboost employs the Rabit~\cite{rabit} communication framework, and LightGBM utilizes the network socket. Here, We use Rabit in Spark GBUN. Rabit is a decentralized distributed communication framework, and we cannot reduce data to a central node by Rabit. Therefore, in each boosting, we execute the ``All Reduce'' operation to obtain the normalized statistical information, the $\mathbf{A}$ and $\mathbf{B}$ matrices for each computing node, and then each node calculate the normalization parameters and solves of the model predicting scores $\mathbf{W}$ by itself. 

Secondly, to further optimize the forwarding efficiency of sparse data, we change the training data storage format from by row to by column. In the by column format, we call the hash function $Km$ times, where $K$ is the number of output neurons, and $m$ is the number of features. For the row storage, we should call the hash function $ K \ cdot nnz $ times. Since the number of features is smaller than nnz in typical cases, the column format reduces forwarding cost. This method is not applied to Python implementation because it causes many random memory accesses and slows down the GPU speed.

Thirdly, we employ Breeze~\cite{breeze} to execute the matrix operations in the Spark implementation. Breeze is a library for numerical processing based on scala. Breeze adopts native \emph{Basic Linear Algebra Subprograms} libraries to speed up the matrix multiplication. However, some pure scala operations, such as summing a matrix and broadcasting a vector to a matrix, are inefficient for some circumstances. We implement our functions for these operations considering the matrix is stored by row or by column, and make the memory access as continue as possible.

Besides, we use the same random seed setting strategy of Python GBUN for Spark GBUN. The only difference is that Spark GBUN uses Breeze to generate a random weight matrix for an untrained neural network. We also take 0 as the default Breeze random seed.

\section{More Details about Experiments}
\begin{table*}[ht]
    \centering
    \caption{Time Costs with Standard Deviation in Accuracy Tests.}\label{tb:data}
    \begin{tabular}{ |l|c|c|c|}
     \hline
     Data
      & XGB & LGB & GBUN \\
     \hline
     Allstate & 21.86 $\pm$ 0.55 &\textbf{2.57$\pm$0.88}&36.72$\pm$4.41\\ 
     \hline
     Epsilon & 9.01$\pm$1.21&\textbf{0.54$\pm$0.06}&0.84$\pm$0.11\\
     \hline
     KDD12 & 218.44$\pm$3.78& \textbf{159.78$\pm$26.47}& 247.56$\pm$4.99\\
     \hline
     News20.b & 1.01$\pm$0.05& \textbf{0.55$\pm$0.06}& 0.79$\pm$0.16\\
     \hline
     Rcv1.b & \textbf{0.10$\pm$0.01}& 0.12$\pm$0.03& 2.50$\pm$0.02\\
     \hline
     \hline
     Aloi & 20.36$\pm$8.50& 5.50$\pm$0.51& \textbf{2.73$\pm$0.30}\\
     \hline
     Letter & 0.04$\pm$ 0.01& \textbf{0.03$\pm$0.02}& 0.55$\pm$0.01\\
     \hline
     News20.m & 1.01$\pm$0.05& 1.22$\pm$0.01& \textbf{0.59$\pm$0.02}\\
     \hline
    \end{tabular}
\end{table*}

\begin{table*}[!t]
    \centering
    \caption{Time Costs of the Distributed Experiments on KDD12.sub Dataset.}\label{tb:kdd12sub}
    \begin{tabular}{ |c|c|c|c|c|c|c|}
     \hline
      \multirow{2}{*}{Model Size} &
      \multirow{2}{*}{Algorithm} &
      \multicolumn{5}{c|}{Number of Machine}
      \\
      \cline{3-7} 
     &  & 4 & 8 & 16 & 32 & 64 \\
     \hline
     \multirow{3}{*}{64} & XGB  &  2.09$\pm$0.35 &1.94$\pm$0.38& 1.71$\pm$0.36 & 1.55$\pm$0.35&1.47$\pm$0.35 \\
     \cline{2-7} 
     & LGB  &  1.27 & 1.45 & 1.96 & 2.46 & 2.67 \\
     \cline{2-7} 
     & GBUN  & 0.86 $\pm$ 0.21 & 0.49$\pm$0.41& 0.31$\pm$0.25 &0.21$\pm$0.13 & 0.20$\pm$0.27 \\
     \hline
     \multirow{3}{*}{1024} & XGB  &  7.50$\pm$1.36 &7.84$\pm$2.13& 11.63$\pm$3.52 & 17.13$\pm$7.67& 24.98$\pm$7.64\\
     \cline{2-7} 
     & LGB  &  7.52 &15.59& 29.60 & 35.40 & 39.58\\
     \cline{2-7} 
     & GBUN  &  31.89$\pm$0.49 &16.19$\pm$0.52& 9.18$\pm$1.44 & 5.62$\pm$0.88& 3.77 $\pm$0.69\\
     \hline
    \end{tabular}
\end{table*}

\begin{table*}[!t]
    \centering
    \caption{GBUN Time Costs of the Distributed Experiments on KDD12 Dataset.}\label{tb:kdd12}
    \begin{tabular}{ |c|c|c|c|c|c|c|c|}
     \hline
      \multirow{2}{*}{Model Size} &
      \multicolumn{6}{c|}{Number of Machine}
      \\
      \cline{2-7} 
     & 16 & 32 & 48 & 64 & 80 & 96 \\
     \hline
     64  & 26.46$\pm$8.40 &15.97$\pm$4.53& 11.00$\pm$3.30 & 8.25$\pm$2.60& 7.16$\pm$2.14 & 6.38 $\pm$ 2.01 \\
     \hline
     1024  & 45.77$\pm$8.68 &30.77$\pm$4.26& 23.30$\pm$3.58 & 18.89$\pm$3.21&16.17$\pm$2.40 & 14.29$\pm$2.05 \\
     \hline
    \end{tabular}
\end{table*}

\begin{table*}[!t]
    \centering
    \caption{Time Costs of the Distributed Experiments with Bandwidth Limitations on KDD12.sub Dataset.}\label{tb:bandwidth}
    \begin{tabular}{ |c|c|c|c|c|c|c|c|}
     \hline
      \multirow{2}{*}{Model Size} &
      \multirow{2}{*}{Algorithm} &
      \multicolumn{5}{c|}{Bandwidth}
      \\
      \cline{3-7} 
     &  & 1GB/s & 100MB/s & 10MB/s & 1MB/s & 100KB/s \\
     \hline
     \multirow{3}{*}{64} & XGB  &  1.71$\pm$0.36 & 8.28$\pm$0.77& 54.42$\pm$7.35 & 559.67$\pm$70.75 & 5778.38$\pm$695.86 \\
     \cline{2-7} 
     & LGB  &  1.96 & 2.56 & 2.98 & 9.29 & 114.70 \\
     \cline{2-7} 
     & GBUN  & 0.31$\pm$0.25 &0.29$\pm$0.22& 0.30$\pm$0.26  &0.37$\pm$0.29&1.25$\pm$0.26 \\
     \hline
     \multirow{3}{*}{1024} & XGB & 11.63$\pm$3.52 & 58.68$\pm$16.53 & 195.09$\pm$46.94& 1820.2295$\pm$451.27 &25728.60$\pm$1920.82  \\
     \cline{2-7} 
     & LGB  &  29.61 & 36.42 & 46.90 & 148.13 & 1249.92 \\
     \cline{2-7} 
     & GBUN  &  9.18$\pm$1.44 &9.27$\pm$1.82&10.09$\pm$2.45 & 17.19$\pm$0.39&97.32$\pm$0.58 \\
     \hline
    \end{tabular}
\end{table*}
\subsection{Random Seed Setting}
We use the default random seed setting of GBUN mentioned in the previous section for both accuracy tests and distributed experiments. For dense data(only accuracy tests use dense datasets),  we use the default random seed setting makes the seed of Pytorch to be 0; for sparse data, GBUN hashes the feature id concatenated with the round number, and the $K$ seeds of the hash functions are from 0 to $K-1$. The random seed setting in our experiments make the training processes and the experimental results are reproducible.

\subsection{More Analysis of the Experimental Results}
Due to the limited space of the main paper, we provide more details of the experimental results here.

Table.~\ref{tb:data} records the average per-round boosting time costs with standard deviations of Xgboost, LightGBM, and GBUN in the accuracy tests. All these means and standard deviations of the training time costs are calculated on the 300 boosting rounds for each test. Based on the standard deviations, we can see that the three algorithms' per-round training time costs are stable in general.

In the main paper, we only illustrate the three algorithms' scaling property curves in distributed experiments because of limited space. Here, we record the average per-round boosting time costs for each experiment in Table.~\ref{tb:kdd12sub},\ref{tb:kdd12},\ref{tb:bandwidth}. For Xgboost and GBUN, we also record the standard deviations for all the tests. All these means and standard deviations of the training time costs are calculated on the 100 boosting rounds for each test. Specifically, the test for Xgboost in the distributed experiment with 100KB/s bandwidth that costs about 7 hours per-round, so we just run 10 rounds to calculate the mean and standard deviation to reduce experiment costs. It should be noted that in all these tables, there are no standard deviation data for LightGBM because each boosting time is recorded in the spark executor logs, but we only save the log of spark driver log during the experiments. Unfortunately, we lose the time cost of each boosting of LightGBM and can't obtain the corresponding standard deviation. 

Table.~\ref{tb:kdd12sub} list the average per-round boosting time costs of the three algorithms on KDD12.sub dataset. Under the ideal scaling property assumption, the training time ratios of the clusters with 8, 16, 32, and 64 machines to the cluster with 4 machines should be 0.5, 0.25, 0.125, and 0.0625. We can see the scaling property of GBUN in Table.~\ref{tb:kdd12sub}: When the model size is 64, the average per-round training time ratios of the clusters with 8, 16, 32, and 64 machines to the cluster with 4 machines are 0.572, 0.366, 0.247, and 0.232; when the model size is 1024, these ratios are 0.508, 0.288, 0.176, and 0.118. The scaling property of the GBUN model with 1024 output neurons is better than that with 64 neurons. Some necessary overhead of distributed implementation may account for a larger proportion of the total cost when the model size is small. When the model size of Xgboost is 64, the training time declines slowly. The average per-round training time ratios of the cluster with 64 machines to the cluster with 4 machines is still 0.702. Worse, when the model size is 1024, the time cost rises continuously. The average per-round training time ratio of the cluster with 64 machines to the cluster with 4 machines is 3.33. The LightGBM continuously increases in training time under both model sizes. When the model size is 64 and 1024, the average per-round training time ratios of the cluster with 64 machines to the cluster with 4 machines are 2.096 and 5.263, respectively. Obviously, the scaling property of Spark GBUN has an overwhelming advantage over Xgboost and LightGBM.

KDD12 is a dataset with the largest number of features in the Libsvm Dataset, reaching 54 million features. For the distributed GBDT algorithm, KDD12 will raise huge communication costs. In our experimental environment, Spark Xgboost and LightGBM can't handle KDD12 at all. On the KDD12 dataset, we test the training time of Spark GBUN on clusters with 16, 32, 48, 64, 80, and 96 machines. Under the ideal scaling property assumptions, the training time ratios of the cluster with 32, 48, 64, 80, and 96 machines to the cluster with 16 machines should be 0.5, 0.333, 0.25, 0.2, and 0.167. From Table.~\ref{tb:kdd12}, we can see that when the model size is 64, the average per-round training time ratios of the cluster with 32, 48, 64, 80, and 96 machines to the cluster with 16 machines are 0.603, 0.416, 0.312, 0.271 and 0.241. While the model is 1024, these ratios are 0.672, 0.509, 0.413, 0.353 and 0.312. We can see that the distributed GBUN has decent property on the ultra-high-dimensional sparse dataset.

Table.~\ref{tb:bandwidth} list the average per-round boosting time costs with standard deviations of the three algorithms on KDD12.sub with limited bandwidths. This table shows how the three algorithms' training time costs change as the network bandwidth decreases. For GBUN with a model size of 64, the per-round training time ratios of the clusters with 100MB/s, 10MB/s, 1MB/s, and 100KB/s bandwidths to the cluster with 1GB/s are 0.9, 1.0, 1.2, and 4.0, respectively; when the model size is 1024, these ratios are 1.0, 1.1, 1.9 and 10.6. Evidently, for GBUN, the training time costs are not affected until the network bandwidth is limited to 1MB/s. When the model size of Xgboost is 64, the per-round training time ratios of the clusters with 100MB/s, 10MB/s, and 100KB/s bandwidths to the cluster with 1GB/s bandwidth are 4.8, 31.8, 326.9, and 3374.9; when the model size is 1024, these ratios are 5.0, 16.8, 156.5, and 2211.7. We can see that the training time of Xgboost is sensitive to the network bandwidth. As the network bandwidth drops by orders of magnitude, the training time also increases exponentially. When the model size of LightGBM is 64, the per-round training time ratios of the clusters with 100MB/s, 10MB/s, 1MB/s, and 100KB/s bandwidths to the cluster with 1GB/s bandwidth are 1.3, 1.5, 4.8, and 58.7; when the model size is 1024, these ratios are 1.2, 1.6, 5.0, and 42.2. The training time of LightGBM is not much affected at 100MB/s and 10MB/s. Since the EFB mode reduces the number of features involved in training, the communication overhead also declines. The comparison of the three algorithms shows that when the model size is 64, the training time cost ratios of GBUN to Xgboost and LightGBM range from bandwidth from 1GB/s to 100KB/s are 1:5.4:6.2, 1:28.6:8.8,  1:179.7:9.8, 1:1510.5:25.1 and 1:4613.8:91.6. When the model size is 1024, these ratios are 1:1.3:3.2, 1:6.3:3.9, 1:19.3:4.6, 1:105.9:8.6, and 1:264.4:12.8. From the above results, we can see that the GBUN algorithm has great advantages in efficiency compared to Xgboost and LightGBM for the distributed learning in a low network bandwidth environment.

\end{appendices}

\bibliography{GBUN.bib}

\end{document}